\documentclass[10pt,twocolumn,letterpaper]{article}

\usepackage[pagenumbers]{cvpr} 

\usepackage{multirow}

\definecolor{cvprblue}{rgb}{0.21,0.49,0.74}
\usepackage[pagebackref,breaklinks,colorlinks,allcolors=cvprblue]{hyperref}

\def\paperID{*****} 
\def\confName{CVPR}
\def\confYear{2026}

\title{Rethinking Pan-sharpening: A New Training Process for Full-Resolution Generalization}

\author{Ran Zhang, Wenbo Xu, Fang Jiabin, Yang Qize, Liu Liu\thanks{Corresponding author.}\\
Hefei University of Technology\\
{\tt\small \{2023212219, 2023170714, 2004fjb, 2023212231\}@mail.hfut.edu.cn, liuliu@hfut.edu.cn}
\and
Xuanhua He, Li Xueheng, Ke Cao\\
University of Science and Technology of China\\
{\tt\small \{hexuanhua, lixueheng, caoke200820\}@mail.ustc.edu.cn}
\and
Jie Zhang\\
Hefei Institutes of Physical Science, Chinese Academy of Sciences\\
{\tt\small zhangjie@iim.ac.cn}
}

\begin{document}
\maketitle

\begin{abstract}
The field of pan-sharpening has recently seen a trend towards increasingly large and complex models, often trained on single, specific satellite datasets. This ``one-dataset, one-model'' approach leads to high computational overhead and impractical deployment. More critically, it overlooks a core challenge: poor generalization from reduced-resolution (RR) training to real-world full-resolution (FR) data. In response to this issue, we challenge this paradigm. We introduce a ``multiple-in-one'' training strategy, where a single, compact model is trained simultaneously on three distinct satellite datasets (WV2, WV3, and GF2). Our experiments show the primary benefit of this unified strategy is a significant and universal boost in FR generalization (QNR) across all tested models, directly addressing this overlooked problem. This paradigm also inherently solves the 'one-model-per-dataset' challenge, and we support it with a highly reproducible, dependency-free codebase for true usability. Finally, we propose PanTiny, a lightweight framework designed specifically for this new, robust paradigm. We demonstrate it achieves a superior performance-to-efficiency balance, proving that principled, simple and robust design is more effective than brute-force scaling in this practical setting. Our work advocates for a community-wide shift towards creating efficient, deployable, and truly generalizable models for pan-sharpening. The code is open-sourced at \url{https://github.com/Zirconium233/PanTiny} .
\end{abstract}

\begin{figure}[t]
\centering
\includegraphics[width=\columnwidth]{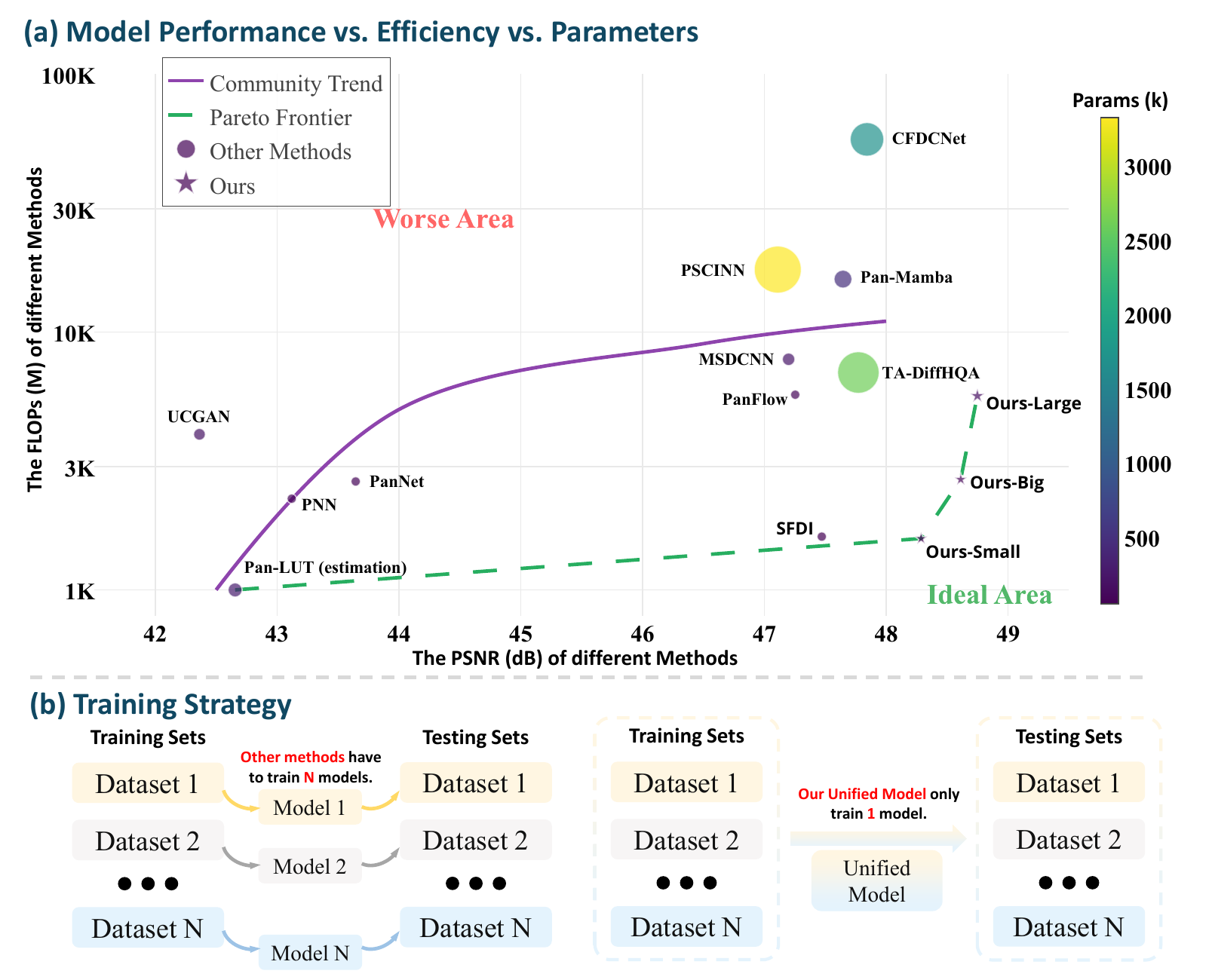}
\caption{Our proposed PanTiny framework enables training a single, unified model on multiple datasets (WV2, WV3, GF2) simultaneously. This all-in-one approach achieves SOTA performance while maintaining a significantly smaller model size and lower computational cost compared to methods that require separate, specialized models for each dataset.}
\label{fig:abstract}
\end{figure}

\section{Introduction}

Pan-sharpening, a fundamental image fusion task in remote sensing, aims to merge a high-resolution panchromatic (PAN) image with a lower-resolution multispectral (LRMS) image to generate a single high-resolution multispectral (HRMS) image \cite{pnn, pannet}. This fused image is crucial for numerous downstream applications \cite{pnn, pannet}. Early approaches were dominated by traditional methods \cite{ihs, pca, king2001wavelet, liu2000smoothing}. The advent of deep learning, particularly with models like PNN \cite{pnn} and PanNet \cite{pannet}, revolutionized the field by learning complex mappings directly from data and significantly improving fusion quality.

However, the recent pursuit of higher performance has led to a research landscape heavily focused on \textit{model architecture innovation}. This trend generally follows three main directions. First, some works attempt to solve cross-dataset generalization, developing complex features to bridge the significant domain gaps between satellites \cite{zhang2024frequency}. Second, a dominant trend is the adaptation of SOTA architectures from other vision fields to pan-sharpening, including porting State-Space Models (SSMs) like Pan-Mamba \cite{panmamba} and flow-based models \cite{yang2023panflownet}. Third, some models achieve high metrics through brute-force scaling, resulting in massive, complex architectures. For instance, CFDCNet \cite{li2025cfdcnet} achieves high metrics but at the cost of an astounding 55G FLOPs.

While these approaches have pushed benchmarks, they share common drawbacks: they are almost all evaluated on reduced-resolution (RR) benchmarks, and they adhere to a ``one-dataset, one-model'' philosophy. We argue this focus on architectural novelty overlooks two more fundamental, practical challenges.
First, the ``one-dataset, one-model'' paradigm is impractical. It requires training and maintaining separate, specialized models for each satellite sensor, complicating real-world deployment.
Second, and most critically, the true, practical challenge is full-resolution (FR) generalization—the ability for a model trained on RR data to perform well on real-world, full-resolution images. This crucial aspect is largely unaddressed. Furthermore, as our supplementary material shows, the domain gaps between satellites are so large that \textit{cross-dataset} generalization (the goal of \cite{zhang2024frequency}) is often futile, suggesting that joint training is a more pragmatic path forward.

In this work, we propose a comprehensive solution that challenges these norms by shifting the focus from \textit{model architecture} to the \textit{training process} and \textit{practical efficiency}.
First, we introduce a new ``all-in-one'' training paradigm, representing a novel process-level innovation. We are the first to demonstrate that a single model can be jointly trained on multiple heterogeneous datasets (WV2, WV3, GF2). The primary benefit is not maximizing RR metrics, but as we will show, dramatically and universally boosting FR generalization (Table~\ref{tab:full_res_wv2}) across all tested models.

Second, this paradigm inherently solves the deployment and usability problem. As illustrated in Figure~\ref{fig:abstract}, a single, compact model can now service three distinct datasets, replacing the cumbersome `one-model-per-dataset' approach.

Finally, we demonstrate that principled, lightweight engineering is superior to brute-force scaling. We propose PanTiny, a lightweight model designed specifically for this paradigm, and a universally powerful composite loss function. Our PanTiny model, benefiting from these innovations, achieves a superior performance-to-efficiency balance, outperforming most larger, specialized models \cite{li2025cfdcnet}. Critically, unlike models such as Pan-Mamba \cite{panmamba} or CFDCNet \cite{li2025cfdcnet}, PanTiny uses only standard, universally available operators, making it dependency-free, easy to reproduce, and highly practical for deployment.

Our contributions are threefold:
\begin{itemize}
    \item We propose and validate a new ``all-in-one'' training paradigm, a process-level innovation that shifts the field's focus from model-centric design and demonstrably boosts generalization on full-resolution data.
    \item We address the `one-model-per-dataset' problem by training a single, unified model for multiple datasets, significantly improving deployment efficiency and usability, as illustrated in Figure~\ref{fig:abstract}.
    \item We propose PanTiny, a lightweight yet powerful model, and a universal composite loss, which together prove that principled, efficient design---which is also dependency-free and easy to reproduce---can surpass brute-force scaling within this new paradigm.
\end{itemize}

\section{Related Work}

\subsection{Traditional and Early Deep Learning Methods}
Traditional pan-sharpening methods include Component Substitution (CS) \cite{ihs, pca}, which enhances spatial detail but risks spectral distortion, and Multi-Resolution Analysis (MRA) \cite{king2001wavelet, liu2000smoothing}, which better preserves spectral information but can add spatial artifacts. Deep learning (DL) methods later surpassed these, with pioneering works like PNN \cite{pnn} using a simple CNN. PanNet \cite{pannet} improved upon this by working in the high-frequency domain, and subsequent models like MSDCNN \cite{msdcnn} explored multi-scale features to enhance fusion quality.

\subsection{SOTA Architectural Innovation}
Recent work is dominated by model-centric innovation, often adapting SOTA architectures from other vision fields. This includes porting State-Space Models (e.g., Pan-Mamba \cite{panmamba}), flow-based models (e.g., PanFlow \cite{yang2023panflownet}), and invertible neural networks (e.g., PSCINN \cite{wang2024panpsci}). Another trend is brute-force scaling, exemplified by CFDCNet \cite{li2025cfdcnet}, which achieves high metrics but at an enormous computational cost (55G FLOPs).

\subsection{Data Generalization Paradigms}
Research on data generalization is also prevalent. Most methods follow a ``one-dataset, one-model'' philosophy. While some attempt \textit{cross-dataset} generalization (e.g., train on A, test on B) by learning domain-irrelevant features \cite{zhang2024frequency}, this approach often fails due to large domain gaps between sensors. The concept of an ``all-in-one'' model, trained jointly on multiple datasets, remains largely unexplored. Our work addresses this gap, proposing a unified paradigm to solve the deployment problem and, critically, to improve \textit{full-resolution generalization}—a practical challenge largely unaddressed by the community.

\begin{figure*}[t]
\centering
\includegraphics[width=\textwidth]{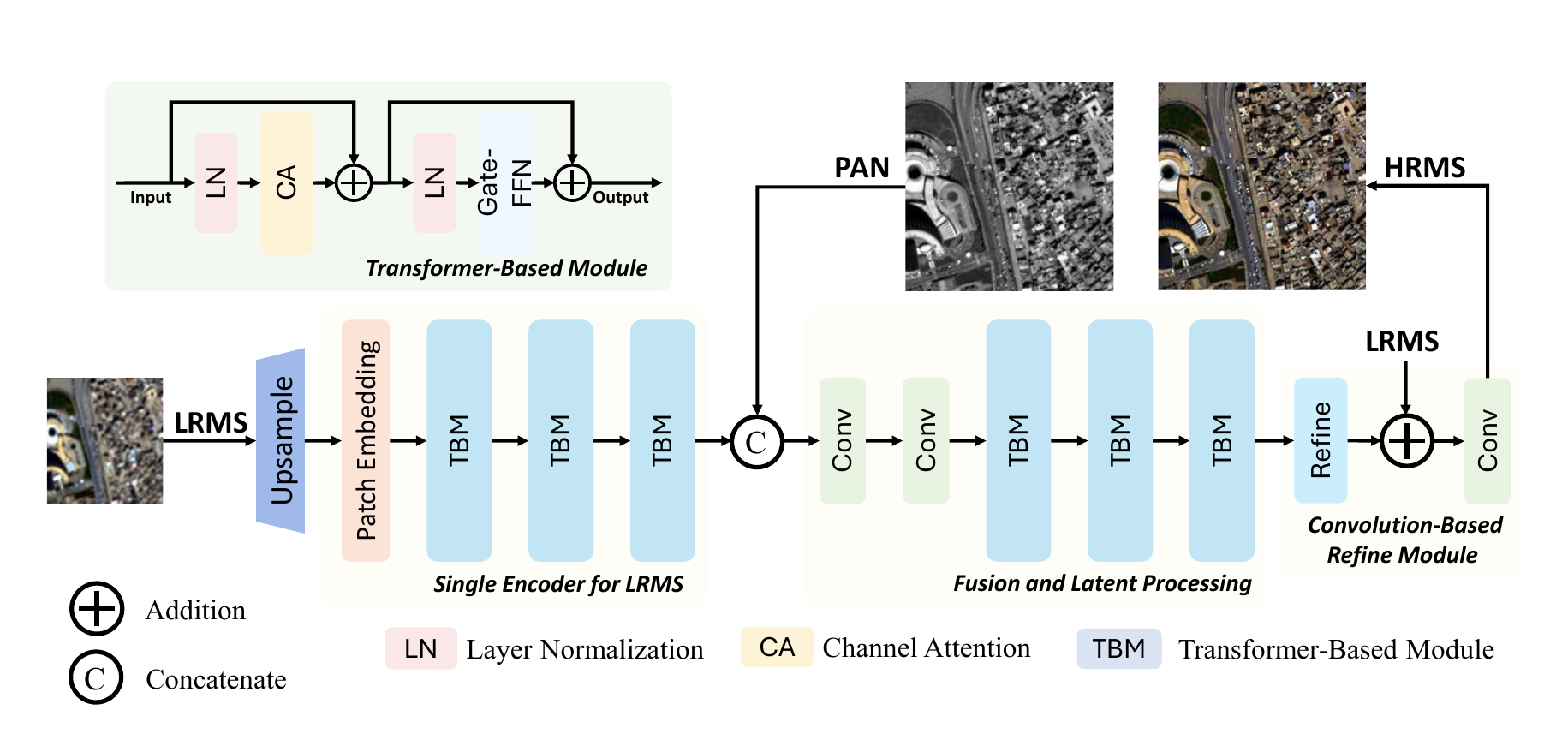}
\caption{The overall architecture of our proposed PanTiny framework. It consists of a single lightweight convolutional encoder for the MS input, a simple yet effective fusion module to integrate PAN information, a body of standard Transformer blocks for deep feature interaction, and a final convolutional layer for refinement.}
\label{fig:pipeline}
\end{figure*}

\section{Methodology}

Our proposed method, PanTiny, is built on the principles of efficiency, simplicity, and empirical validation, designed specifically to be robust and effective within our ``all-in-one'' training paradigm. As our new focus is on multi-domain training, we deliberately avoid overly complex operators that, as we will show (Table~\ref{tab:fusion_ablation}), are prone to overfitting. We instead focus on a clean, effective architecture where each component's inclusion is justified by extensive experiments. The overall architecture, shown in Figure~\ref{fig:pipeline}, features a single-encoder design, a Transformer-based body for feature processing, and a simple convolutional refinement head.

\subsection{Overall Architecture}

\subsubsection{Patch Embedding and Encoding} Unlike many methods that use separate encoders, we adopt an efficient single-encoder architecture. The upsampled MS image (bms) is first passed through an OverlapPatchEmbed layer (a 3x3 convolution) to extract initial features. These features are then processed by a shallow set of Transformer blocks (self.encoder in our implementation).

\subsubsection{Feature Fusion Module} \textbf{Following the initial encoding}, the PAN image is integrated directly in the feature space via our fusion module. Our investigation into fusion mechanisms revealed a surprising insight: \textbf{in the multi-dataset ``all-in-one'' context, simplicity triumphs over complexity.} We found that complex, specialized fusion strategies like the multi-layer ``deepfusion'' block \cite{panmamba} (Table~\ref{tab:fusion_ablation}) \textbf{degraded performance, suggesting overfitting} to dataset-specific artifacts. This "specialized knowledge" fails when the model must perform across multiple domains. Therefore, we adopt a simple but highly effective Enhanced Conv'' block (two consecutive 3x3 convolutional layers), which our ablations (Table~\ref{tab:fusion_ablation}) proved to be the most robust.

\subsubsection{Deep Feature Processing and Refinement} The fused features are then processed by the main body of the network (self.refinement in code), which consists of a deeper stack of standard Transformer blocks for effective long-range dependency modeling. For the final reconstruction, we again prioritize simplicity. Our experiments (Table~\ref{tab:refine_ablation}) confirmed that a \textbf{single convolutional layer}  was optimal, while more elaborate refinement modules offered no significant benefit and unnecessarily increased model size.

\begin{table*}[t]
\centering
\renewcommand{\arraystretch}{1.15} 
\resizebox{\textwidth}{!}{%
\begin{tabular}{l|cc|ccc|ccc|ccc}
\toprule
\multirow{2}{*}{\textbf{Model}} & \multirow{2}{*}{\textbf{Params(K)}} & \multirow{2}{*}{\textbf{FLOPs(G)}} & \multicolumn{3}{c}{\textbf{WV2}} & \multicolumn{3}{c}{\textbf{WV3}} & \multicolumn{3}{c}{\textbf{GF2}} \\
\cmidrule(lr){4-6} \cmidrule(lr){7-9} \cmidrule(lr){10-12}
& & & \textbf{PSNR}$\uparrow$ & \textbf{SSIM}$\uparrow$ & \textbf{SAM}$\downarrow$ & \textbf{PSNR}$\uparrow$ & \textbf{SSIM}$\uparrow$ & \textbf{SAM}$\downarrow$ & \textbf{PSNR}$\uparrow$ & \textbf{SSIM}$\uparrow$ & \textbf{SAM}$\downarrow$ \\
\midrule
\multicolumn{12}{c}{\textit{Traditional Methods}} \\
\midrule
Brovey \cite{brovey} & - & - & 35.86 & 0.9216 & 0.0403 & 22.50 & 0.5466 & 0.1159 & 37.79 & 0.9026 & 0.0218 \\
IHS \cite{ihs} & - & - & 35.29 & 0.9027 & 0.0461 & 22.55 & 0.5354 & 0.1266 & 38.17 & 0.9100 & 0.0243 \\
SFIM \cite{liu2000smoothing} & - & - & 34.12 & 0.8975 & 0.0439 & 21.82 & 0.5457 & 0.1208 & 36.90 & 0.8882 & 0.0318 \\
GS \cite{GS} & - & - & 35.63 & 0.9176 & 0.0423 & 22.56 & 0.5470 & 0.1217 & 37.22 & 0.9034 & 0.0309 \\
\midrule
\multicolumn{12}{c}{\textit{Deep Learning Methods (All-in-One Training)}} \\
\midrule
PNN \cite{pnn} & 68.9 & 2.26 & 39.82 & 0.9540 & 0.0282 & 29.49 & 0.9005 & 0.0861 & 43.14 & 0.9667 & 0.0178 \\
PanNet \cite{pannet} & 80.3 & 2.63 & 38.98 & 0.9468 & 0.0301 & 29.12 & 0.8927 & 0.0935 & 43.26 & 0.9668 & 0.0176 \\
MSDCNN \cite{msdcnn} & 239.0 & 7.83 & 40.31 & 0.9580 & 0.0267 & 29.63 & 0.9033 & 0.0833 & 43.21 & 0.9671 & 0.0176 \\
PanFlow \cite{yang2023panflownet} & 87.3 & 2.86 & 41.11 & 0.9645 & 0.0243 & 30.04 & 0.9106 & 0.0799 & 46.36 & 0.9825 & 0.0125 \\
PSCINN \cite{wang2024panpsci} & 3321.5 & 108.84 & 35.60 & 0.8967 & 0.0336 & 22.61 & 0.5538 & 0.1115 & 42.69 & 0.9616 & 0.0181 \\
Pan-Mamba \cite{panmamba} & 488.8 & 16.02 & 41.39 & 0.9663 & 0.0236 & 30.17 & 0.9174 & 0.0779 & 43.98 & 0.9725 & 0.0164 \\
CFDCNet \cite{li2025cfdcnet} & 1700.8 & 55.73 & 41.54 & 0.9667 & 0.0233 & \underline{30.42} & 0.9155 & 0.0775 & 47.76 & \underline{0.9866} & 0.0107 \\
\midrule
PanTiny (Small) & \textbf{48.3} & \textbf{1.58} & \underline{41.62} & \underline{0.9685} & \underline{0.0230} & 30.38 & \underline{0.9216} & \underline{0.0768} & \underline{48.16} & \textbf{0.9884} & \underline{0.0099} \\
\textbf{PanTiny (Big)} & 81.7 & 2.68 & \textbf{41.85} & \textbf{0.9696} & \textbf{0.0224} & \textbf{30.59} & \textbf{0.9238} & \textbf{0.0749} & \textbf{48.61} & \textbf{0.9894} & \textbf{0.0095} \\
\bottomrule
\end{tabular}%
}
\caption{Main quantitative comparison. All models are trained simultaneously on WV2, WV3, and GF2 datasets and evaluated on each using a single model. Our PanTiny (Big) achieves the best performance across all datasets. `-' indicates a traditional, non-learning based approach. Best results are in \textbf{bold}, second-best are \underline{underlined}.}
\label{tab:main_comparison}
\end{table*}

\subsection{Transformer Block}
While the overall structure is inspired by the original Transformer, our implementation uses a Pre-LayerNorm (Pre-LN) configuration for improved training stability. For an input feature map $X_{l-1}$, the output $X_l$ of a single Transformer block is computed as:
\begin{align}
    X'_{l} &= \text{CA}(\text{LN}(X_{l-1})) + X_{l-1} \\
    X_{l} &= \text{GDFN}(\text{LN}(X'_{l})) + X'_{l}
\end{align}
where LN denotes Layer Normalization, CA is our Channel Attention module, and GDFN is a Gated-DConv Feed-Forward Network.

\paragraph{Channel Attention (CA).}
The CA module captures global context by performing self-attention across channel dimensions. Given an input $X \in \mathbb{R}^{B \times C \times H \times W}$, we first generate the query ($Q$), key ($K$), and value ($V$) projections via depth-wise convolutions. The attention map is then computed as:
\begin{equation}
\text{Attention}(Q, K, V) = \text{Softmax}( (Q_n K_n^T) \cdot \tau)V_n
\end{equation}
where $Q_n$ and $K_n$ are L2-normalized query and key tensors, and $\tau$ is a learnable temperature parameter that scales the attention map. This design avoids the standard scaling by feature dimension, instead allowing the network to learn the optimal attention scaling.

\paragraph{Gated-DConv Feed-Forward Network (GDFN).}
To enhance feature representation efficiently, we employ a gated feed-forward network. An input tensor is first projected to a higher-dimensional space and then split into two parallel paths, $X_1$ and $X_2$. The output is computed as:
\begin{equation}
\text{GDFN}(X) = \text{Conv}_{\text{out}}(\text{GELU}(\text{DWConv}(X_1)) \odot \text{DWConv}(X_2))
\end{equation}
where $\odot$ denotes element-wise multiplication. This gating mechanism allows for more dynamic and expressive feature transformations compared to a standard FFN.

\begin{figure*}[h]
\centering
\includegraphics[width=\textwidth]{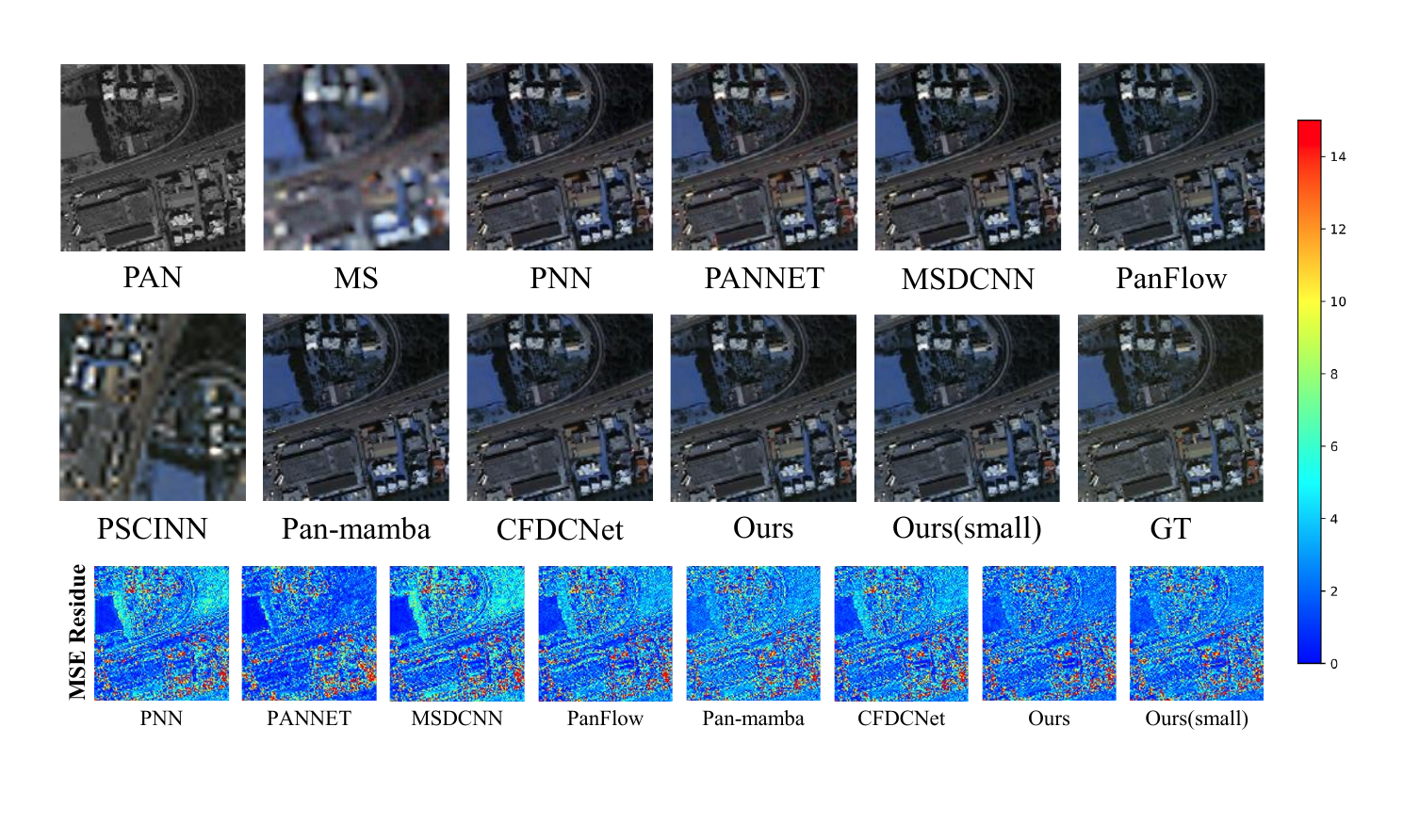}
\caption{Quality comparison across SOTA methods on WV3 dataset. Refer to supplementary materials for more results. }
\label{fig:vp}
\end{figure*}

\begin{table}[h]
\centering
\resizebox{\columnwidth}{!}{%
\begin{tabular}{l|c|c}
\toprule
\multirow{2}{*}{\textbf{Model}} & \multicolumn{2}{c}{\textbf{QNR} $\uparrow$} \\
\cmidrule(lr){2-3}
& \textbf{Separate} & \textbf{All-in-One} \\
\midrule
MSDCNN \cite{msdcnn} & 0.7683 & \underline{0.8898} \\
PanNet \cite{pannet} & 0.7684 & 0.8726 \\
PNN \cite{pnn} & 0.7527 & 0.8844 \\
PSCINN \cite{wang2024panpsci} & 0.7904 & 0.8849 \\
PanTiny (Small) & 0.7827 & 0.8751 \\
PanTiny (Big) & 0.7985 & 0.8793 \\
PanFlow \cite{yang2023panflownet} & 0.7910 & \textbf{0.8900} \\
\bottomrule
\end{tabular}%
}
\caption{Generalization on WV2 full-resolution data. All-in-one training significantly boosts the QNR metric for all models.}
\label{tab:full_res_wv2}
\end{table}

\subsection{Loss Function}
The choice of loss function is critical for training high-performance restoration models. While many prior works rely solely on the L1 loss, our empirical study showed that a composite loss function yields substantially better results. Our total loss $L_{total}$ is a weighted sum of three components, applied to the model's output $O$ and the ground truth $G$:
\begin{equation}
L_{total} = \lambda_1 L_{1} + \lambda_2 L_{SSIM} + \lambda_3 L_{Focal}
\end{equation}

\begin{itemize}
    \item \textbf{L1 Loss}: We use the Charbonnier loss \cite{charbonnier1994two}, a differentiable variant of the L1 norm that is less sensitive to outliers. For a batch of $B$ images with $N$ pixels each, it is defined as:
    \begin{equation}
        L_1 = \frac{1}{B \cdot N} \sum_{i=1}^{B \cdot N} \sqrt{(O_i - G_i)^2 + \epsilon^2}
    \end{equation}
    where $\epsilon$ is a small constant (e.g., $10^{-6}$) for numerical stability.

    \item \textbf{SSIM Loss}: To preserve perceptual quality and high-frequency structural details, we incorporate the Structural Similarity (SSIM) loss \cite{wang2004image}. The SSIM index between two image patches $o$ and $g$ is:
    \begin{equation}
        \text{SSIM}(o, g) = \frac{(2\mu_o\mu_g + C_1)(2\sigma_{og} + C_2)}{(\mu_o^2 + \mu_g^2 + C_1)(\sigma_o^2 + \sigma_g^2 + C_2)}
    \end{equation}
    where $\mu$ and $\sigma$ represent the mean and variance, and $C_1, C_2$ are stabilizing constants. The final loss is computed as $L_{SSIM} = 1 - \text{SSIM}(O, G)$, averaged over all patches. Our ablations in Table~\ref{tab:loss_ablation} clearly show its importance.
    
    \item \textbf{Focal-Inspired Regression Loss}: Inspired by the \textit{principle} of focal loss in classification, we adapt its core idea to regression to prioritize ``hard'' pixels. We term this $L_{Focal}$. Let $d_i = |O_i - G_i|$ be the absolute error for a given pixel $i$ (assumed to be in the [0, 1] normalized range). Our loss is formulated as: \begin{equation} L_{Focal} = \frac{1}{B \cdot N} \sum_{i=1}^{B \cdot N} \frac{(255 \cdot d_i)^{r_1}}{255} \cdot d_i \end{equation} where the constant $255$ scales the normalized error back to the 8-bit dynamic range before applying the exponent, stabilizing the gradient. \textbf{In our experiments, we use a focusing parameter $r_1 = 1.1$.} This formulation acts as a non-linear weighting function: by setting $r_1 > 1.0$, pixels with larger errors (hard'' pixels) are polynomially up-weighted, compelling the model to focus on challenging details, similar in effect to a high-order $L_p$ loss.
\end{itemize}
Through extensive experiments (detailed in Table~\ref{tab:loss_ablation} and our supplementary material), we determined the optimal weights to be $\lambda_1=1.5$, $\lambda_2=4.0$, and $\lambda_3=1.5$, which consistently delivered the best performance.

\section{Experiments}

\begin{table*}[t]
\centering
\resizebox{\textwidth}{!}{%
\begin{tabular}{l|l|ccc|ccc|ccc}
\toprule
\multirow{2}{*}{\textbf{Model}} & \multirow{2}{*}{\textbf{Training}} & \multicolumn{3}{c}{\textbf{WV2}} & \multicolumn{3}{c}{\textbf{WV3}} & \multicolumn{3}{c}{\textbf{GF2}} \\
\cmidrule(lr){3-5} \cmidrule(lr){6-8} \cmidrule(lr){9-11}
& & \textbf{PSNR}$\uparrow$ & \textbf{SSIM}$\uparrow$ & \textbf{SAM}$\downarrow$ & \textbf{PSNR}$\uparrow$ & \textbf{SSIM}$\uparrow$ & \textbf{SAM}$\downarrow$ & \textbf{PSNR}$\uparrow$ & \textbf{SSIM}$\uparrow$ & \textbf{SAM}$\downarrow$ \\
\midrule
\multirow{2}{*}{Pan-Mamba \cite{panmamba}} & All-in-One & 41.39 & 0.9663 & 0.0236 & 30.17 & 0.9174 & 0.0779 & 43.98 & 0.9725 & 0.0164 \\
& Separate & \textbf{42.24} & \textbf{0.9729} & \textbf{0.0212} & \textbf{31.16} & \textbf{0.9299} & \textbf{0.0702} & \textbf{47.65} & \textbf{0.9894} & \textbf{0.0103} \\
\midrule
\multirow{2}{*}{PNN \cite{pnn}} & All-in-One & 39.82 & 0.9540 & 0.0282 & 29.49 & 0.9005 & 0.0861 & \textbf{43.14} & 0.9667 & \textbf{0.0178} \\
& Separate & \textbf{40.76} & \textbf{0.9624} & \textbf{0.0259} & \textbf{29.94} & \textbf{0.9121} & \textbf{0.0824} & 43.12 & \textbf{0.9704} & \underline{0.0172} \\
\midrule
\multirow{2}{*}{PanFlow \cite{yang2023panflownet}} & All-in-One & 41.11 & 0.9645 & 0.0243 & 30.04 & 0.9106 & 0.0799 & 46.36 & 0.9825 & 0.0125 \\
& Separate & \textbf{41.86} & \textbf{0.9712} & \textbf{0.0224} & \textbf{30.49} & \textbf{0.9221} & \textbf{0.0751} & \textbf{47.25} & \textbf{0.9884} & \textbf{0.0103} \\
\midrule
\multirow{2}{*}{PSCINN \cite{wang2024panpsci}} & All-in-One & 35.60 & 0.8967 & 0.0336 & 22.61 & 0.5538 & 0.1115 & 42.69 & 0.9616 & 0.0181 \\
& Separate & \textbf{41.85} & \textbf{0.9703} & \textbf{0.0223} & \textbf{30.56} & \textbf{0.9230} & \textbf{0.0748} & \textbf{47.11} & \textbf{0.9878} & \textbf{0.0107} \\
\midrule
\multirow{2}{*}{CFDCNet \cite{li2025cfdcnet}} & All-in-One & 41.54 & 0.9667 & 0.0233 & 30.42 & 0.9155 & 0.0775 & 47.76 & 0.9866 & 0.0107 \\
& Separate & \textbf{42.24} & \textbf{0.9733} & \textbf{0.0209} & \textbf{31.24} & \textbf{0.9327} & \textbf{0.0694} & \textbf{47.84} & \textbf{0.9902} & \textbf{0.0097} \\
\midrule
\multirow{2}{*}{Pan-LUT \cite{cai2025panlut}} & All-in-One & - & - & - & - & - & - & - & - & - \\
& Separate & \textbf{39.84} & \textbf{0.9555} & \textbf{0.0286} & \textbf{28.82} & \textbf{0.8936} & \textbf{0.0935} & \textbf{42.66} & \textbf{0.9642} & \textbf{0.0189} \\
\midrule
\multirow{2}{*}{\textbf{Ours (PanTiny Big)}} & All-in-One & \underline{41.85} & \underline{0.9696} & \underline{0.0224} & \underline{30.59} & \underline{0.9238} & \underline{0.0749} & \underline{48.61} & \underline{0.9894} & \underline{0.0095} \\
& Separate & \textbf{42.16} & \textbf{0.9711} & \textbf{0.0217} & \textbf{30.61} & \textbf{0.9245} & \textbf{0.0747} & \textbf{48.93} & \textbf{0.9900} & \textbf{0.0092} \\
\bottomrule
\end{tabular}%
}
\caption{Performance comparison between ``all-in-one'' and ``separate'' dataset training. ``Separate'' results are from original papers. The performance gap highlights the generalization challenge for complex, specialized models.}
\label{tab:all_vs_separate}
\end{table*}

\subsection{Setup}
\textbf{Datasets.} We conduct experiments on three public datasets: WorldView-2 (WV2), WorldView-3 (WV3), and GaoFen-2 (GF2). \textbf{It is a standard community practice \cite{pnn, pannet} to pre-process the 8-band WV2 and WV3 data to 4-bands (R, G, B, NIR) to match GF2. We follow this standard, using the 4-band versions} to ensure fair comparison and enable our ``all-in-one'' joint training. For our primary ``all-in-one'' experiments, we combine the training sets of all three. We follow standard protocols for evaluation, using both reduced-resolution and full-resolution test sets. 

\textbf{Evaluation Metrics.} We provide a comprehensive evaluation using both reference and no-reference metrics. For reduced-resolution evaluation, we use Peak Signal-to-Noise Ratio (PSNR), Structural Similarity (SSIM) \cite{wang2004image}, Spectral Angle Mapper (SAM) \cite{sam}, and ERGAS \cite{ergas}. For full-resolution evaluation, we use the no-reference metrics D$_\lambda$, D$_s$, and QNR \cite{qnr}.

\textbf{Implementation Details.} Our framework is implemented in PyTorch. All models are trained on a single NVIDIA RTX 4090 GPU. We use the ADAM optimizer with a learning rate of $5 \times 10^{-4}$ and betas of $(0.9, 0.999)$. A cosine annealing scheduler adjusts the learning rate over 500 epochs with a batch size of 16.

\subsection{The ``All-in-One'' Training Paradigm}
Our experimental evaluation is first designed to validate our core thesis: that the ``all-in-one'' training paradigm is a more robust and effective method for developing generalizable pan-sharpening models, solving the key challenges of FR generalization and deployment practicality.

\begin{figure}[h]
\centering
\includegraphics[width=\columnwidth]{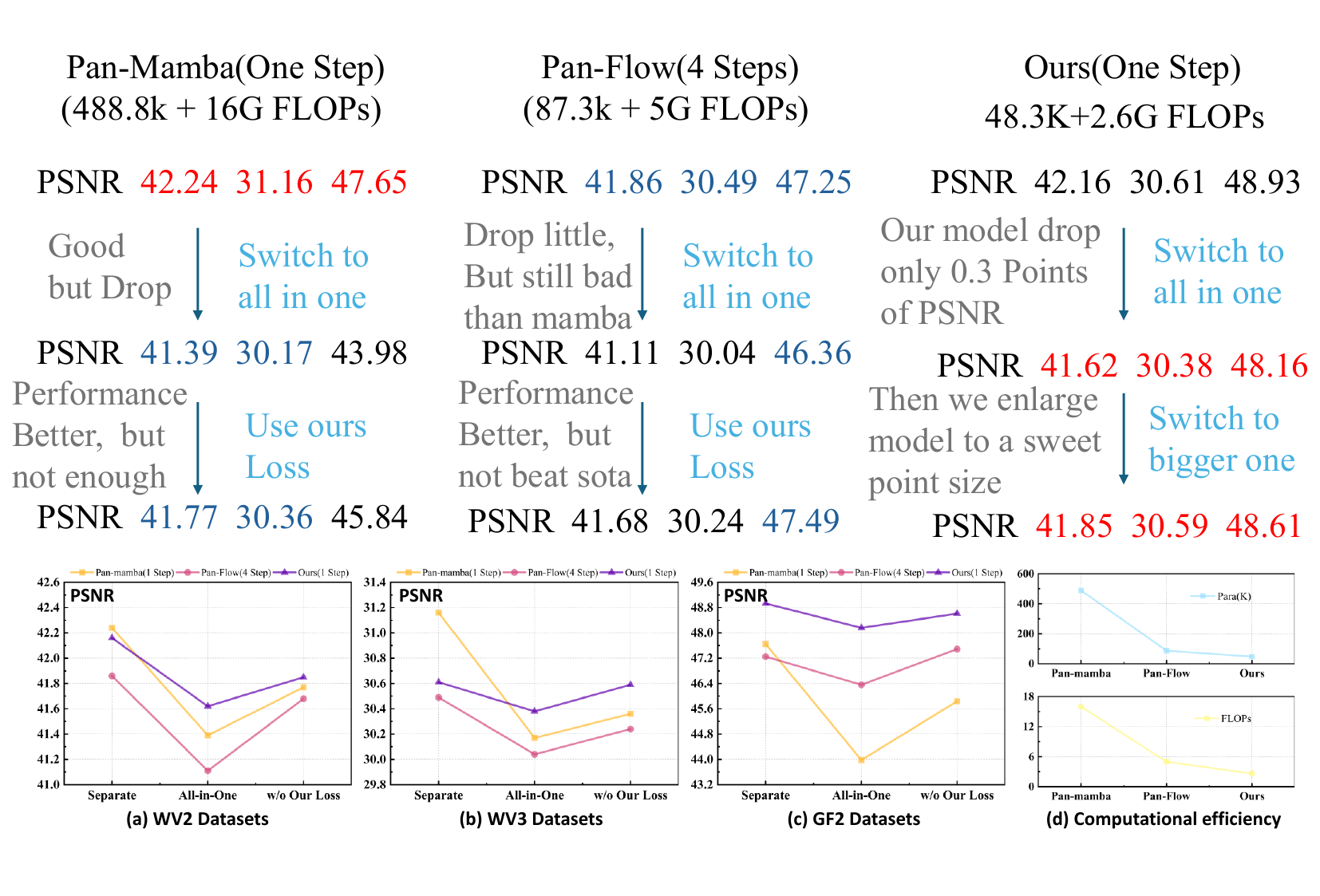}
\caption{Performance trajectory of the ablation study. This figure illustrates the performance changes of different models under the ``all-in-one'' training paradigm.}
\label{fig:ablation_trajectory}
\end{figure}

\begin{table*}[t]
\centering
\resizebox{\textwidth}{!}{%
\begin{tabular}{l|cc|ccc|ccc|ccc}
\toprule
\multirow{2}{*}{\textbf{Model}} & \multirow{2}{*}{\textbf{Params(K)}} & \multirow{2}{*}{\textbf{FLOPs(G)}} & \multicolumn{3}{c}{\textbf{WV2}} & \multicolumn{3}{c}{\textbf{WV3}} & \multicolumn{3}{c}{\textbf{GF2}} \\
\cmidrule(lr){4-6} \cmidrule(lr){7-9} \cmidrule(lr){10-12}
& & & \textbf{PSNR}$\uparrow$ & \textbf{SSIM}$\uparrow$ & \textbf{SAM}$\downarrow$ & \textbf{PSNR}$\uparrow$ & \textbf{SSIM}$\uparrow$ & \textbf{SAM}$\downarrow$ & \textbf{PSNR}$\uparrow$ & \textbf{SSIM}$\uparrow$ & \textbf{SAM}$\downarrow$ \\
\midrule
CFDCNet \cite{li2025cfdcnet} & 1700.8 & 55.73 & \textbf{42.50} & \textbf{0.9729} & \textbf{0.0205} & \textbf{31.11} & \textbf{0.9298} & \textbf{0.0707} & \textbf{49.07} & \textbf{0.9903} & \textbf{0.0091} \\
Pan-Mamba \cite{panmamba} & 488.8 & 16.02 & 41.77 & 0.9691 & 0.0226 & 30.36 & 0.9215 & 0.0769 & 45.84 & 0.9811 & 0.0134 \\
PanFlow \cite{yang2023panflownet} & 87.3 & 2.86 & 41.68 & 0.9688 & 0.0229 & 30.24 & 0.9197 & 0.0785 & 47.49 & 0.9865 & 0.0109 \\
MSDCNN \cite{msdcnn} & 239.0 & 7.83 & 41.46 & 0.9669 & 0.0236 & 30.18 & 0.9189 & 0.0786 & 44.09 & 0.9730 & 0.0161 \\
PNN \cite{pnn} & 68.9 & 2.26 & 40.84 & 0.9635 & 0.0256 & 29.82 & 0.9128 & 0.0834 & 43.40 & 0.9688 & 0.0173 \\
PanNet \cite{pannet} & 80.3 & 2.63 & 40.79 & 0.9620 & 0.0256 & 29.82 & 0.9106 & 0.0841 & 43.76 & 0.9705 & 0.0167 \\
\midrule
DeepPNN (ours) & 271.1 & 8.88 & 41.89 & 0.9700 & 0.0224 & 30.43 & 0.9228 & 0.0759 & 47.45 & 0.9869 & 0.0109 \\
ResAtten (ours) & 263.0 & 8.62 & \underline{41.97} & \underline{0.9702} & \underline{0.0222} & 30.40 & 0.9223 & 0.0759 & 47.14 & 0.9856 & 0.0113 \\
\textbf{PanTiny (Big)} & \textbf{81.7} & \textbf{2.68} & 41.85 & 0.9696 & 0.0224 & \underline{30.59} & \underline{0.9238} & \underline{0.0749} & \underline{48.61} & \underline{0.9894} & \underline{0.0095} \\
PanTiny (Small) & 48.3 & 1.58 & 41.62 & 0.9685 & 0.0230 & 30.38 & 0.9216 & 0.0768 & 48.16 & 0.9884 & 0.0099 \\
\bottomrule
\end{tabular}%
}
\caption{Model architecture ablation under our unified loss. Our proposed loss benefits all models, but our architecture remains highly competitive. Performance gains over original reported results are due to our improved training strategy.}
\label{tab:model_abl}
\end{table*}

\subsubsection{Limitations of Traditional Paradigms}
We first analyze the impact of our ``all-in-one'' paradigm on the traditional reduced-resolution (RR) benchmarks, as shown in Table~\ref{tab:all_vs_separate}. The results are mixed: compared to ``Separate'' models that are highly specialized (overfitted) to a single dataset, the ``All-in-One'' models (e.g., Pan-Mamba \cite{panmamba} and PSCINN \cite{wang2024panpsci}) show a significant performance drop on RR metrics. This highlights that the domain gaps between satellites are severe, and models must balance conflicting objectives. In contrast, our ``PanTiny (Big)'' model is remarkably robust, with only a minor drop on WV2/GF2 and almost no change on WV3. This demonstrates that simply chasing RR benchmark scores on a single dataset is not a robust strategy.

\subsubsection{The True Value: Full-Resolution Generalization}
The primary goal of pan-sharpening is real-world application, which requires generalizing from RR training to FR inference. Table~\ref{tab:full_res_wv2} reveals the true, unambiguous value of our paradigm. For \textbf{all models tested}, switching from ``Separate'' to ``All-in-One'' training results in a \textbf{dramatic and universal improvement in the full-resolution QNR metric}. For instance, PanFlow's \cite{yang2023panflownet} QNR score on WV2 jumps from 0.7910 to 0.8900, and MSDCNN's \cite{msdcnn} from 0.7683 to 0.8898. This is clear evidence that our paradigm trains models to be far more robust and generalizable to the real-world FR task. This, combined with the inherent benefit of deploying a single unified model (our Contribution 2), underscores the power of our proposed method.

\subsection{PanTiny: An Efficient SOTA Model for the New Paradigm}
Having established the ``all-in-one'' paradigm as a superior and more practical approach, we now benchmark models within this framework to validate our proposed PanTiny (Contribution 3).

\subsubsection{Main Results on Multi-Dataset Training}
Table~\ref{tab:main_comparison} presents the main results. Our proposed PanTiny (Big) achieves SOTA performance across all three datasets, outperforming both classic and recent methods. Notably, it surpasses CFDCNet \cite{li2025cfdcnet}, a much larger model, on all metrics. PanTiny (Small) also achieves a highly competitive performance-to-efficiency balance. The results for PSCINN \cite{wang2024panpsci} highlight its instability in this multi-dataset setting, as it failed to complete training, which we consider a practical failure, further validating our robust design choice.

\begin{table}[ht]
\centering
\resizebox{\columnwidth}{!}{%
\begin{tabular}{l|c|cc|cc|cc}
\toprule
\multirow{2}{*}{\textbf{Model}} & \textbf{Params} & \multicolumn{2}{c|}{\textbf{WV2}} & \multicolumn{2}{c|}{\textbf{WV3}} & \multicolumn{2}{c}{\textbf{GF2}} \\
& \textbf{(K)} & \textbf{PSNR} & \textbf{SSIM} & \textbf{PSNR} & \textbf{SSIM} & \textbf{PSNR} & \textbf{SSIM} \\
\midrule
PanTiny (Small) & 48.3 & 41.62 & 0.9685 & 30.38 & 0.9216 & 48.16 & 0.9884 \\
PanTiny (Big) & 81.7 & 41.85 & 0.9696 & 30.59 & 0.9238 & 48.61 & 0.9894 \\
PanTiny (Large Body) & 172.4 & \underline{42.06} & \underline{0.9708} & \underline{30.67} & \underline{0.9248} & \underline{48.75} & \underline{0.9896} \\
PanTiny (Huge Body) & 195.9 & \textbf{42.12} & \textbf{0.9711} & \textbf{30.74} & \textbf{0.9258} & \textbf{48.85} & \textbf{0.9898} \\
\bottomrule
\end{tabular}%
}
\caption{Ablation on model size. Brute-force scaling yields diminishing returns compared to our efficient `Big' design.}
\label{tab:scaling_ablation}
\end{table}

\subsubsection{Architectural Design and Efficiency}
PanTiny's SOTA performance is achieved with superior efficiency. As shown in Table~\ref{tab:main_comparison}, its 81.7K parameters and 2.68G FLOPs are orders of magnitude smaller than heavyweight competitors like CFDCNet (1700.8K params, 55.73G FLOPs) \cite{li2025cfdcnet}.
We further explore scaling in Table~\ref{tab:scaling_ablation}. By expanding our model to a 'Huge' version (195.9K params), we can nearly match CFDCNet's best-in-class performance on GF2 (48.85 vs 49.07 PSNR) at less than 12\% of its parameter count. This reinforces our core argument that principled, efficient engineering surpasses brute-force scaling.
Critically, PanTiny's practical efficiency also lies in its \textbf{reproducibility}. Unlike Pan-Mamba \cite{panmamba} or CFDCNet \cite{li2025cfdcnet} which may rely on specialized CUDA kernels or dependencies, PanTiny is built with standard PyTorch operators, ensuring it is easy to deploy and validate.

\begin{table*}[t]
\centering
\resizebox{\textwidth}{!}{%
\begin{tabular}{l|ccc|ccc|ccc}
\toprule
\multirow{2}{*}{\textbf{Loss Combination (L1, SSIM, Focal)}} & \multicolumn{3}{c}{\textbf{WV2}} & \multicolumn{3}{c}{\textbf{WV3}} & \multicolumn{3}{c}{\textbf{GF2}} \\
\cmidrule(lr){2-4} \cmidrule(lr){5-7} \cmidrule(lr){8-10}
 & \textbf{PSNR}$\uparrow$ & \textbf{SSIM}$\uparrow$ & \textbf{SAM}$\downarrow$ & \textbf{PSNR}$\uparrow$ & \textbf{SSIM}$\uparrow$ & \textbf{SAM}$\downarrow$ & \textbf{PSNR}$\uparrow$ & \textbf{SSIM}$\uparrow$ & \textbf{SAM}$\downarrow$ \\
\midrule
L1 only (1.0, 0, 0) & 39.77 & 0.9532 & 0.0285 & 29.19 & 0.8939 & 0.0953 & 45.42 & 0.9782 & 0.0141 \\
SSIM only (0, 1.0, 0) & 40.82 & 0.9648 & 0.0254 & 29.91 & 0.9158 & 0.0816 & 47.21 & 0.9865 & 0.0111 \\
Focal only (0, 0, 1.0) & 39.87 & 0.9545 & 0.0281 & 29.18 & 0.8937 & 0.0933 & 44.80 & 0.9757 & 0.0150 \\
\midrule
Balanced (0.8, 0.5, 0.4) & 41.00 & 0.9640 & 0.0248 & 29.99 & 0.9128 & 0.0832 & 47.38 & 0.9859 & 0.0110 \\
Equal (1.0, 1.0, 1.0) & 41.28 & 0.9659 & 0.0240 & 30.17 & 0.9170 & 0.0791 & 47.68 & 0.9869 & 0.0105 \\
High Weight (3.0, 3.0, 3.0) & \underline{41.68} & 0.9683 & \underline{0.0229} & \textbf{30.49} & 0.9214 & 0.0762 & \textbf{48.30} & \underline{0.9885} & \underline{0.0099} \\
SSIM Focus (0.5, 8.0, 0.5) & \underline{41.68} & \textbf{0.9694} & \textbf{0.0228} & 30.45 & \textbf{0.9233} & \textbf{0.0760} & 48.17 & \textbf{0.9887} & 0.0100 \\
\textbf{Ours (1.5, 4.0, 1.5)} & \textbf{41.70} & \underline{0.9689} & \textbf{0.0228} & \underline{30.46} & \underline{0.9225} & \underline{0.0761} & \underline{48.29} & \textbf{0.9887} & \textbf{0.0098} \\
\bottomrule
\end{tabular}%
}
\caption{Ablation study on loss function components and weights using our PanTiny model. Our proposed combination (1.5, 4.0, 1.5) provides the best overall performance.}
\label{tab:loss_ablation}
\end{table*}

\subsection{Ablation Studies and Further Analysis}
We validate the specific design choices for our loss function and the PanTiny architecture.

\subsubsection{PanTiny's Architectural Ablations}
We now validate the internal design choices of PanTiny to demonstrate that its simplicity is a principled and effective strategy. We conducted detailed ablations on its core fusion and refinement components.

\textbf{Fusion Module:} The fusion mechanism is critical for integrating PAN and MS information. As shown in Table~\ref{tab:fusion_ablation}, we tested several alternatives. A simple ``1x1 Conv'' provides a strong baseline, but our proposed ``Enhanced Conv'' (a 2-layer Conv) achieves the best overall performance. Notably, we also tested the complex ``DeepFusion'' module from Pan-Mamba \cite{panmamba}. This module, while effective in its original single-dataset setting, performs poorly here. This result strongly suggests that complex, specialized modules tend to overfit to dataset-specific artifacts, making them less suitable for the robust, multi-dataset "all-in-one" paradigm. This validates our choice of a simpler, more robust block.

\textbf{Refinement Module:} We applied the same principle to the refinement block, as shown in Table~\ref{tab:refine_ablation}. Our final model uses a simple ``Conv'' layer. We found that adding complexity, such as a ``Large Conv'' or a ``Channel Attn.'' block, unnecessarily increases the parameter count without yielding a consistent or significant performance benefit. These ablations on both the fusion and refinement modules confirm that PanTiny's simplicity is not naive, but rather a deliberate and empirically-validated design choice that prioritizes robustness and parameter efficiency in our unified training framework.

\begin{table}[ht]
\centering
\resizebox{\columnwidth}{!}{%
\begin{tabular}{l|c|ccc}
\toprule
\textbf{Fusion Type} & \textbf{Params(K)} & \textbf{WV2} & \textbf{WV3} & \textbf{GF2} \\
\midrule
1x1 Conv & 68.2 & \underline{41.75} & \underline{30.45} & \underline{48.37} \\
Channel Attn. & 71.4 & 41.72 & 30.44 & 48.34 \\
Gated Conv & 70.3 & 41.66 & 30.44 & 48.32 \\
DeepFusion \cite{panmamba} & 113.6 & 41.66 & 30.34 & 48.35 \\
\textbf{Enhanced Conv (Ours)} & \textbf{81.7} & \textbf{41.85} & \textbf{30.59} & \textbf{48.61} \\
\bottomrule
\end{tabular}%
}
\caption{Ablation on the fusion module. (Showing only PSNR).}
\label{tab:fusion_ablation}
\end{table}

\begin{table}[h]
\centering
\resizebox{\columnwidth}{!}{%
\begin{tabular}{l|c|ccc}
\toprule
\textbf{Refine Type} & \textbf{Params(K)} & \textbf{WV2} & \textbf{WV3} & \textbf{GF2} \\
\midrule
\textbf{Conv (Ours)} & \textbf{81.7} & \textbf{41.90} & \textbf{30.61} & 48.49 \\
Channel Attn. & 96.4 & \textbf{41.90} & \underline{30.55} & \underline{48.50} \\
Large Conv & 88.8 & \underline{41.87} & 30.49 & \textbf{48.52} \\
\bottomrule
\end{tabular}%
}
\caption{Ablation on the refinement module. (Showing only PSNR).}
\label{tab:refine_ablation}
\end{table}

\subsubsection{Universality of the Composite Loss}
To demonstrate the superiority of our loss function as a general tool, we conducted a benchmark where all competing methods were trained with it (Table~\ref{tab:model_abl}). The pan-sharpening community has historically relied heavily on the standard L1 loss, often used in isolation, as it is simple and stable. However, our initial ablations in Table~\ref{tab:loss_ablation} show that using a single loss component, including L1, yields suboptimal performance.

This motivated an extensive, systematic \textbf{loss search} to find an optimal composite weighting. Our final weights ($\lambda_1=1.5, \lambda_2=4.0, \lambda_3=1.5$) are the result of this search, which is detailed further in our supplementary material. A key insight from this process was that while L1 provides a stable regression base, assigning a \textbf{particularly high weight to the SSIM component} ($\lambda_2=4.0$) consistently delivered the best performance across datasets, especially for GF2. As shown in Table~\ref{tab:model_abl}, when competing models (often trained with a simple L1 loss in a `Separate' paradigm) are retrained with our unified strategy and this optimized composite loss, they see a significant performance uplift. This demonstrates our composite loss's power as a general tool for the community.

\section{Conclusion}
In this paper, we challenged the pan-sharpening community's prevailing focus on model-centric innovation and complex cross-dataset generalization. We shifted the focus to two more practical, yet overlooked, challenges: full-resolution (FR) generalization and deployment usability. Our primary contribution is the novel ``all-in-one'' training paradigm, which we proved is a universally effective strategy for dramatically boosting the FR generalization (e.g., QNR) of all tested models. This paradigm also inherently solves the ``one-model-per-dataset'' problem, and we support this with a highly reproducible, dependency-free codebase for true usability. Finally, we proposed PanTiny, a lightweight model designed for this new paradigm. We demonstrated that this simple, efficient architecture is the SOTA performer in this practical setting, proving that principled, robust design is more effective than the complex architectural scaling seen in other works. We believe our combined contributions—the paradigm for FR generalization, the focus on usability, and the efficient PanTiny model and a universal composite loss—offer a more sustainable and practical path forward for pan-sharpening research. 





{
    \small
    \bibliographystyle{ieeenat_fullname}
    \bibliography{main}
}

\clearpage
\setcounter{page}{1}
\maketitlesupplementary

\section{On the Limitations of Generalization and the Necessity of All-in-One Training}
\label{sec:supp_generalization}

In the main paper, we argue that our ``all-in-one'' training paradigm is a superior approach to achieving robust pan-sharpening models compared to existing generalization methods. Here, we provide detailed experimental evidence to support this claim. The conventional approach to generalization—training a model on a single source dataset and testing it on multiple unseen target datasets—often fails to bridge the significant domain gap between different satellite sensors. We contend that this approach often leads to models that are merely overfitted to the source domain, rather than being truly generalizable.

\subsection{The Challenge of Domain Gaps in Pan-sharpening Datasets}
A fundamental challenge in pan-sharpening is the significant domain gap between datasets from different satellite sensors. For instance, the datasets used in our study—WorldView-2 (WV2), WorldView-3 (WV3), and GaoFen-2 (GF2)—exhibit substantial differences. WV2 and WV3 provide 8-band multispectral images, which are conventionally processed down to 4 bands for standard pan-sharpening tasks, whereas GF2 directly provides 4-band data. Furthermore, these satellites operate with different sensors, at different altitudes, and capture images with varying ground resolutions and atmospheric conditions. This inherent data heterogeneity means that a model optimized for one dataset's specific spectral and spatial characteristics will inevitably struggle to perform well on another. This large domain gap makes true generalization exceptionally difficult and underscores the limitations of single-dataset training.

\subsection{The ``One-Epoch Generalization'' Illusion}
To test the hypothesis of overfitting in conventional generalization studies, we conducted a surprising experiment: we trained several of our intermediate models for only \textbf{one epoch} on the WV2 dataset and then evaluated their performance on the unseen WV3 and GF2 datasets. The results, shown in Table~\ref{tab:one_epoch}, are striking. When tested on GF2, our one-epoch trained ``M4 (Channel Attn)'' model achieves a PSNR of 39.13. This result is comparable to or even surpasses the performance of fully-trained models from dedicated generalization papers, such as DDIF \cite{chen2022learning}, which reports a PSNR of 37.77 on GF2 after being fully trained on WV2 (see Table~\ref{tab:from_ddif_no_ergas}). This suggests that the hundreds of additional training epochs in those works contribute little to true generalization, instead primarily reinforcing the model's bias towards the source dataset. This finding strongly motivates a shift away from the ``train-on-one, test-on-many'' methodology. Any reviewer can easily verify this conclusion with a personal computer in under 5 minutes using our provided codebase, if they already have the datasets.

\begin{table*}[t]
\centering
\resizebox{\textwidth}{!}{%
\begin{tabular}{l|ccc|ccc|ccc}
\toprule
\multirow{2}{*}{\textbf{Method}} & \multicolumn{3}{c|}{\textbf{WorldView-III}} & \multicolumn{3}{c|}{\textbf{Worldview-II}} & \multicolumn{3}{c}{\textbf{GaoFen2}} \\
\cmidrule(lr){2-4} \cmidrule(lr){5-7} \cmidrule(lr){8-10}
& \textbf{PSNR}$\uparrow$ & \textbf{SSIM}$\uparrow$ & \textbf{SAM}$\downarrow$ & \textbf{PSNR}$\uparrow$ & \textbf{SSIM}$\uparrow$ & \textbf{SAM}$\downarrow$ & \textbf{PSNR}$\uparrow$ & \textbf{SSIM}$\uparrow$ & \textbf{SAM}$\downarrow$ \\
\midrule
PNN & 21.9204 & 0.5771 & 0.1301 & 40.8487 & 0.9642 & 0.0254 & 28.6188 & 0.8649 & 0.1177 \\
PANNET & 22.3157 & 0.5597 & 0.1273 & 40.8176 & 0.9626 & 0.0257 & 35.0812 & 0.8707 & 0.0422 \\
MSDCNN & 21.2841 & 0.5651 & 0.1551 & 41.3355 & 0.9664 & 0.0242 & 29.6255 & 0.8815 & 0.1062 \\
DICNN & 19.1958 & 0.5606 & 0.1453 & 39.9554 & 0.9597 & 0.0275 & 34.4568 & \underline{0.8857} & 0.0447 \\
SRPPNN & 22.0543 & 0.5779 & 0.1340 & 41.4538 & 0.9679 & 0.0233 & 33.7282 & 0.7989 & 0.0513 \\
Panformer & 19.3288 & 0.5715 & 0.1533 & 41.2170 & 0.9672 & 0.0239 & 23.4309 & 0.8192 & 0.2239 \\
Mutual & 21.7467 & 0.5783 & 0.1488 & 41.6773 & 0.9705 & 0.0224 & 34.0899 & 0.8380 & 0.0523 \\
LAGConv & 21.6249 & 0.5520 & 0.1516 & 41.6815 & 0.9598 & 0.0325 & 35.1923 & 0.8753 & 0.0436 \\
SFIIN & 21.9983 & 0.5766 & 0.1310 & 41.7080 & 0.9693 & 0.0228 & \underline{36.7285} & 0.8705 & \underline{0.0307} \\
P2Net & \underline{22.4445} & \underline{0.6084} & \underline{0.1258} & \textbf{41.9229} & \underline{0.9711} & \underline{0.0219} & 35.4512 & 0.8383 & 0.0386 \\
\midrule
\textbf{DDIF} & \textbf{22.9937} & \textbf{0.6102} & \textbf{0.1213} & \underline{41.7219} & \textbf{0.9719} & \textbf{0.0217} & \textbf{37.7663} & \textbf{0.8919} & \textbf{0.0253} \\
\bottomrule
\end{tabular}%
}
\caption{Quantitative comparison from a prior generalization work \cite{chen2022learning}, with the model trained on the Worldview-II dataset and tested on other datasets. The best results are marked in \textbf{bold} and the second results are marked with \underline{underline}. $\uparrow$ indicates that the larger the value, the better the performance, and $\downarrow$ indicates that the smaller the value, the better the performance.}
\label{tab:from_ddif_no_ergas}
\end{table*}

\begin{table*}[h!]
\centering
\resizebox{\textwidth}{!}{%
\begin{tabular}{l|cc|ccc|ccc|ccc}
\toprule
\multirow{2}{*}{\textbf{Model}} & \multirow{2}{*}{\textbf{Params(K)}} & \multirow{2}{*}{\textbf{FLOPs(G)}} & \multicolumn{3}{c}{\textbf{WV2}} & \multicolumn{3}{c}{\textbf{WV3}} & \multicolumn{3}{c}{\textbf{GF2}} \\
\cmidrule(lr){4-6} \cmidrule(lr){7-9} \cmidrule(lr){10-12}
& & & \textbf{PSNR}$\uparrow$ & \textbf{SSIM}$\uparrow$ & \textbf{SAM}$\downarrow$ & \textbf{PSNR}$\uparrow$ & \textbf{SSIM}$\uparrow$ & \textbf{SAM}$\downarrow$ & \textbf{PSNR}$\uparrow$ & \textbf{SSIM}$\uparrow$ & \textbf{SAM}$\downarrow$ \\
\midrule
M3 (Dual Enc.) & 128.9 & 4.22 & 38.00 & 0.9347 & 0.0352 & 22.12 & 0.5608 & 0.1272 & 35.90 & 0.9227 & 0.0378 \\
pantiny(small) (Single Enc.) & 48.3 & 1.58 & 36.38 & 0.9127 & 0.0396 & 22.05 & 0.5352 & 0.1317 & 34.39 & 0.9230 & 0.0627 \\
M4 (Gated Conv) & 67.0 & 2.20 & 36.84 & 0.9195 & 0.0358 & 22.09 & 0.5668 & 0.1264 & 37.67 & 0.9450 & 0.0297 \\
M4 (Channel Attn) & 66.0 & 2.16 & 36.87 & 0.9174 & 0.0358 & 22.34 & 0.5710 & 0.1243 & 39.13 & 0.9263 & 0.0260 \\
\bottomrule
\end{tabular}%
}
\caption{Performance of various intermediate models after only \textbf{one epoch} of training on the WV2 dataset, tested on all three datasets. The competitive results on unseen domains (WV3, GF2) challenge the effectiveness of conventional generalization strategies.}
\label{tab:one_epoch}
\end{table*}

\subsection{The Overfitting Trap of Separate Training}
Further evidence against the separate training paradigm comes from analyzing the cross-domain performance of our own model when fully trained on a single dataset. Table~\ref{tab:separate_overfitting} shows the results of training ``PanTiny (Big)'' to convergence on one source dataset and testing on all three. For instance, the model trained on WV2 achieves an excellent 42.16 PSNR on its own test set, but its performance plummets to 21.76 on WV3 and 33.92 on GF2. This performance is substantially worse than the one-epoch results, proving that prolonged training on a single dataset actively harms its ability to generalize by causing it to overfit to the source domain's specific characteristics.

\begin{table*}[h!]
\centering
\resizebox{\textwidth}{!}{%
\begin{tabular}{l|ccc|ccc|ccc}
\toprule
\multirow{2}{*}{\textbf{Training Dataset}} & \multicolumn{3}{c}{\textbf{Test on WV2}} & \multicolumn{3}{c}{\textbf{Test on WV3}} & \multicolumn{3}{c}{\textbf{Test on GF2}} \\
\cmidrule(lr){2-4} \cmidrule(lr){5-7} \cmidrule(lr){8-10}
& \textbf{PSNR}$\uparrow$ & \textbf{SSIM}$\uparrow$ & \textbf{SAM}$\downarrow$ & \textbf{PSNR}$\uparrow$ & \textbf{SSIM}$\uparrow$ & \textbf{SAM}$\downarrow$ & \textbf{PSNR}$\uparrow$ & \textbf{SSIM}$\uparrow$ & \textbf{SAM}$\downarrow$ \\
\midrule
WV2 Only & \textbf{42.16} & \textbf{0.9711} & \textbf{0.0217} & 21.76 & 0.5628 & 0.1284 & 33.92 & 0.8899 & 0.0433 \\
WV3 Only & 27.99 & 0.7880 & 0.0964 & \textbf{30.61} & \textbf{0.9245} & \textbf{0.0747} & 24.75 & 0.6798 & 0.0863 \\
GF2 Only & 34.40 & 0.8882 & 0.0438 & 21.89 & 0.4650 & 0.1279 & \textbf{48.93} & \textbf{0.9900} & \textbf{0.0092} \\
\bottomrule
\end{tabular}%
}
\caption{Cross-domain performance of ``PanTiny (Big)'' when trained separately on a single source dataset. The drastic performance drop on target datasets highlights the overfitting issue inherent in this paradigm.}
\label{tab:separate_overfitting}
\end{table*}

\subsection{Failure Case: Generalization to Jilin-1 Dataset}
To push the boundaries of generalization, we tested our all-in-one trained models on the Jilin-1 dataset, which was completely unseen during training. As shown in Table~\ref{tab:jilin_test}, the performance of all models is poor, indicating that even our robust ``all-in-one'' paradigm has its limits when faced with a significant domain shift. Interestingly, PSCINN, which performed poorly on the training datasets, shows relatively better (though still low) performance here, possibly due to its different architectural inductive biases. This experiment reinforces our central thesis: true generalization in pan-sharpening is a data problem, and robust performance requires training on diverse, representative datasets.

\begin{table}[h!]
\centering
\resizebox{0.8\columnwidth}{!}{%
\begin{tabular}{l|ccc}
\toprule
\multirow{2}{*}{\textbf{Model}} & \multicolumn{3}{c}{\textbf{Jilin-1}} \\
\cmidrule(lr){2-4}
& \textbf{PSNR}$\uparrow$ & \textbf{SSIM}$\uparrow$ & \textbf{SAM}$\downarrow$ \\
\midrule
PNN & 22.16 & 0.6000 & 0.1286 \\
PanNet & 22.82 & \underline{0.6255} & 0.0911 \\
PanFlow & 22.14 & 0.5641 & \underline{0.0861} \\
MSDCNN & 21.73 & 0.5988 & 0.1321 \\
PSCINN & \textbf{27.90} & \textbf{0.8319} & \textbf{0.0812} \\
Ours (PanTiny Big) & \underline{23.10} & 0.5694 & 0.0884 \\
\bottomrule
\end{tabular}%
}
\caption{Zero-shot generalization performance on the unseen Jilin-1 dataset. All models were trained under the ``all-in-one'' paradigm. The best results are in \textbf{bold} and the second results are marked with \underline{underline}.}
\label{tab:jilin_test}
\end{table}

\section{Detailed Ablation on the Composite Loss}
\label{sec:supp_loss}

\subsection{The Overlooked Potential of Loss Functions}
Historically, the pan-sharpening community has predominantly focused on advancing model architectures to achieve performance gains. The L1 loss has long been the de-facto standard, with the majority of research efforts dedicated to designing more sophisticated networks. However, this model-centric approach appears to be reaching a point of diminishing returns. As evidenced by recent SOTA models like CFDCNet \cite{li2025cfdcnet}, achieving marginal performance improvements now requires an enormous increase in computational cost (over 55G FLOPs), suggesting an architectural bottleneck.

We posit that the loss function, a relatively underexplored area, holds the key to unlocking the next level of performance. While perceptual losses like SSIM \cite{wang2004image} have been considered, they were often dismissed after preliminary tests showed that using them in isolation or with balanced weights did not yield superior results and could sometimes introduce color artifacts. This led to a widespread underestimation of their potential. We believe that a systematic, large-scale exploration of loss combinations has been a missing piece in the field.

\subsection{Our Systematic Two-Stage Search for the Optimal Loss}
Our work is the first, to our knowledge, to conduct such an extensive search. This process, detailed in Table~\ref{tab:loss_full_ablation}, was divided into two stages.

In the first stage, we conducted a broad search using our ``PanTiny (Big)'' model to understand the general behavior of different loss component weightings. We tested balanced configurations like (1,1,1) as well as configurations focusing on each individual component. This initial exploration yielded a crucial insight: combinations with a high weight on the SSIM component, such as (1,3,1), consistently outperformed others.

Guided by this finding, we initiated a second, more fine-grained search stage. To accelerate experimentation, we used our lighter ``PanTiny (Small)'' model and focused exclusively on high-SSIM weight combinations. This meticulous process allowed us to identify the ``(1.5, 4.0, 1.5)'' configuration as the most robust and highest-performing combination. This discovery is not just a set of tuned hyperparameters; it represents a universally applicable principle that can elevate the entire field. By applying this composite loss, we have unlocked a new tier of performance, pushing the SOTA for metrics like GF2 PSNR into the 48-49 dB era for a wide range of models.

\begin{table*}[h!]
\centering
\resizebox{\textwidth}{!}{%
\begin{tabular}{l|l|ccc|ccc|ccc}
\toprule
\multirow{2}{*}{\textbf{Loss Combination (L1, SSIM, Focal)}} & \multirow{2}{*}{\textbf{Model}} & \multicolumn{3}{c}{\textbf{WV2}} & \multicolumn{3}{c}{\textbf{WV3}} & \multicolumn{3}{c}{\textbf{GF2}} \\
\cmidrule(lr){3-5} \cmidrule(lr){6-8} \cmidrule(lr){9-11}
 & & \textbf{PSNR}$\uparrow$ & \textbf{SSIM}$\uparrow$ & \textbf{SAM}$\downarrow$ & \textbf{PSNR}$\uparrow$ & \textbf{SSIM}$\uparrow$ & \textbf{SAM}$\downarrow$ & \textbf{PSNR}$\uparrow$ & \textbf{SSIM}$\uparrow$ & \textbf{SAM}$\downarrow$ \\
\midrule
\multicolumn{11}{c}{\textit{Stage 1: Broad Search on PanTiny (Big)}} \\
\midrule
L1 only (1.0, 0, 0) & pantiny & 39.77 & 0.9532 & 0.0285 & 29.19 & 0.8939 & 0.0953 & 45.42 & 0.9782 & 0.0141 \\
SSIM only (0, 1.0, 0) & pantiny & 40.82 & 0.9648 & 0.0254 & 29.91 & 0.9158 & 0.0816 & 47.21 & 0.9865 & 0.0111 \\
Focal only (0, 0, 1.0) & pantiny & 39.87 & 0.9545 & 0.0281 & 29.18 & 0.8937 & 0.0933 & 44.80 & 0.9757 & 0.0150 \\
Balanced (0.8, 0.5, 0.4) & pantiny & 41.00 & 0.9640 & 0.0248 & 29.99 & 0.9128 & 0.0832 & 47.38 & 0.9859 & 0.0110 \\
Equal (1.0, 1.0, 1.0) & pantiny & 41.28 & 0.9659 & 0.0240 & 30.17 & 0.9170 & 0.0791 & 47.68 & 0.9869 & 0.0105 \\
SSIM Focus (1.0, 3.0, 1.0) & pantiny & \textbf{41.57} & \textbf{0.9680} & \textbf{0.0232} & \textbf{30.37} & \textbf{0.9213} & \textbf{0.0771} & \textbf{48.14} & \textbf{0.9882} & \textbf{0.0100} \\
L1 Focus (3.0, 1.0, 1.0) & pantiny & 41.38 & 0.9663 & 0.0237 & \underline{30.31} & \underline{0.9186} & \underline{0.0777} & \underline{47.99} & \underline{0.9877} & \underline{0.0102} \\
Focal Focus (1.0, 1.0, 3.0) & pantiny & \underline{41.41} & \underline{0.9665} & \underline{0.0236} & 30.28 & 0.9177 & \underline{0.0777} & 47.92 & 0.9875 & \underline{0.0102} \\
\midrule
\multicolumn{11}{c}{\textit{Stage 2: Fine-grained Search on PanTiny (Small)}} \\
\midrule
(2.0, 2.0, 2.0) & panrestormer & 41.60 & 0.9680 & 0.0231 & 30.38 & 0.9206 & 0.0768 & 48.17 & 0.9884 & \underline{0.0099} \\
(3.0, 0.8, 1.0) & panrestormer & 41.32 & 0.9661 & 0.0237 & 30.30 & 0.9180 & 0.0777 & 47.95 & 0.9876 & 0.0102 \\
(0.8, 0.8, 3.0) & panrestormer & 41.35 & 0.9664 & 0.0237 & 30.27 & 0.9177 & 0.0779 & 48.13 & 0.9880 & \underline{0.0099} \\
(0.8, 5.0, 1.0) & panrestormer & 41.66 & \underline{0.9689} & \textbf{0.0228} & 30.40 & \underline{0.9227} & 0.0767 & 48.25 & \textbf{0.9887} & \underline{0.0099} \\
(1.5, 3.5, 1.5) & panrestormer & 41.64 & 0.9686 & \underline{0.0229} & 30.42 & 0.9219 & 0.0765 & \underline{48.28} & \underline{0.9886} & \textbf{0.0098} \\
(0.8, 3.0, 1.0) & panrestormer & 41.52 & 0.9681 & 0.0232 & 30.39 & 0.9213 & 0.0768 & 48.06 & 0.9883 & 0.0101 \\
(0.5, 8.0, 0.5) & panrestormer & \underline{41.68} & \textbf{0.9694} & \textbf{0.0228} & \underline{30.45} & \textbf{0.9233} & \textbf{0.0760} & 48.17 & \textbf{0.9887} & 0.0100 \\
\textbf{(1.5, 4.0, 1.5)} & panrestormer & \textbf{41.70} & \underline{0.9689} & \textbf{0.0228} & \textbf{30.46} & 0.9225 & \underline{0.0761} & \textbf{48.29} & \textbf{0.9887} & \textbf{0.0098} \\
\bottomrule
\end{tabular}%
}
\caption{Full ablation study on loss function components and weights. The top part shows a broad search on our ``PanTiny (Big)'' model, while the bottom part shows a fine-grained search on the ``PanTiny (Small)'' model to accelerate experiments. Our proposed combination (1.5, 4.0, 1.5) provides the best overall performance. Best results are in \textbf{bold}, second-best are \underline{underlined}.}
\label{tab:loss_full_ablation}
\end{table*}

\section{Detailed Ablation on Model Architecture}
\label{sec:supp_arch}
Our final PanTiny architecture was the result of a systematic exploration of different design choices, moving from complex structures to a refined, efficient final model. Our initial explorations included models with multiple downsampling levels and dual-encoder designs (named M3, M4, M5), but these were ultimately superseded by the more efficient single-encoder architecture of PanTiny.

\subsection{Downsampling Strategy}
A common strategy in image restoration is to use a U-Net-like architecture with multiple downsampling stages to capture multi-scale features. We investigated this by creating variants of our base model (``PanTiny(Small)'') with 0, 2, and 4 downsampling levels, using a basic L1 loss for a fair architectural comparison. As shown in Table~\ref{tab:downsample_abl}, we found that increasing the downsampling levels led to a significant increase in parameters and a decrease in overall performance. The 0-level model (no downsampling) performed the best, indicating that for pan-sharpening, maintaining the full feature resolution is more effective. This led us to adopt a flat, single-scale architecture for PanTiny.

\begin{table*}[h!]
\centering
\resizebox{\textwidth}{!}{%
\begin{tabular}{l|cc|ccc|ccc|ccc}
\toprule
\multirow{2}{*}{\textbf{Model}} & \multirow{2}{*}{\textbf{Params(K)}} & \multirow{2}{*}{\textbf{FLOPs(G)}} & \multicolumn{3}{c}{\textbf{WV2}} & \multicolumn{3}{c}{\textbf{WV3}} & \multicolumn{3}{c}{\textbf{GF2}} \\
\cmidrule(lr){4-6} \cmidrule(lr){7-9} \cmidrule(lr){10-12}
& & & \textbf{PSNR}$\uparrow$ & \textbf{SSIM}$\uparrow$ & \textbf{SAM}$\downarrow$ & \textbf{PSNR}$\uparrow$ & \textbf{SSIM}$\uparrow$ & \textbf{SAM}$\downarrow$ & \textbf{PSNR}$\uparrow$ & \textbf{SSIM}$\uparrow$ & \textbf{SAM}$\downarrow$ \\
\midrule
4-ds & 446.7 & 14.64 & 39.59 & 0.9543 & 0.0287 & \underline{28.93} & 0.8926 & 0.0977 & 45.33 & 0.9791 & \underline{0.0140} \\
2-ds & 121.2 & 3.97 & \textbf{40.74} & \textbf{0.9627} & \textbf{0.0255} & \textbf{29.58} & \underline{0.9079} & \underline{0.0856} & \textbf{46.74} & \textbf{0.9840} & \textbf{0.0118} \\
\textbf{0-ds (Ours)} & \textbf{48.0} & \textbf{1.57} & \underline{40.58} & \underline{0.9618} & \underline{0.0257} & \textbf{29.58} & \textbf{0.9083} & \textbf{0.0849} & \underline{46.64} & \underline{0.9839} & \textbf{0.0118} \\
\bottomrule
\end{tabular}%
}
\caption{Ablation on downsampling levels using a simple L1 loss. Deeper U-Net-like structures did not improve performance. Best results are in \textbf{bold}, second-best are \underline{underlined}.}
\label{tab:downsample_abl}
\end{table*}

\subsection{Investigating the ``DeepFusion'' Module}
In our main paper, we noted that Pan-Mamba's performance degrades significantly in the ``all-in-one'' setting. We hypothesized this was due to its complex ``DeepFusion'' module overfitting to single-dataset characteristics. To verify this, we integrated the ``DeepFusion'' block into our ``m6'' experimental model. As shown in Table~\ref{tab:deepfusion_abl}, not only does the ``DeepFusion'' block increase parameter count, but it also consistently underperforms compared to simpler fusion mechanisms like our ``Enhanced Conv'' (from the main paper's ablation) or even basic ``Gated Conv'' and ``Channel Attention''. Furthermore, increasing the depth of the ``DeepFusion'' block from 2 to 5 layers leads to a further drop in performance. This provides strong evidence that such complex fusion modules, while effective for a single dataset, are detrimental to generalization in the ``all-in-one'' paradigm.

\begin{table*}[h!]
\centering
\resizebox{\textwidth}{!}{%
\begin{tabular}{l|c|ccc|ccc|ccc}
\toprule
\multirow{2}{*}{\textbf{Fusion Type}} & \multirow{2}{*}{\textbf{Params(K)}} & \multicolumn{3}{c}{\textbf{WV2}} & \multicolumn{3}{c}{\textbf{WV3}} & \multicolumn{3}{c}{\textbf{GF2}} \\
\cmidrule(lr){3-5} \cmidrule(lr){6-8} \cmidrule(lr){9-11}
& & \textbf{PSNR}$\uparrow$ & \textbf{SSIM}$\uparrow$ & \textbf{SAM}$\downarrow$ & \textbf{PSNR}$\uparrow$ & \textbf{SSIM}$\uparrow$ & \textbf{SAM}$\downarrow$ & \textbf{PSNR}$\uparrow$ & \textbf{SSIM}$\uparrow$ & \textbf{SAM}$\downarrow$ \\
\midrule
Gated Conv & \textbf{63.2} & \underline{40.95} & \underline{0.9639} & \textbf{0.0248} & \textbf{30.04} & \textbf{0.9149} & \textbf{0.0800} & \underline{47.37} & \underline{0.9860} & \textbf{0.0108} \\
Channel Attention & 64.3 & \textbf{41.06} & \textbf{0.9641} & \textbf{0.0248} & \underline{30.00} & \underline{0.9133} & 0.0830 & \textbf{47.54} & \textbf{0.9864} & \textbf{0.0108} \\
DeepFusion (2 layers) & 75.2 & 40.78 & 0.9625 & \underline{0.0254} & 29.88 & 0.9127 & \underline{0.0822} & 46.99 & 0.9851 & \underline{0.0113} \\
DeepFusion (5 layers) & 106.5 & 40.67 & 0.9625 & 0.0255 & 29.87 & 0.9122 & 0.0823 & 46.87 & 0.9848 & 0.0116 \\
\bottomrule
\end{tabular}%
}
\caption{Ablation on the ``DeepFusion'' module using our ``m6'' variant. Complex, deep fusion strategies underperform simpler ones in the multi-dataset setting.}
\label{tab:deepfusion_abl}
\end{table*}

\subsection{Single-Encoder vs. Dual-Encoder Design}
In our architectural exploration, we also compared single-encoder and dual-encoder designs. Our ``m5'' model variant features a dual-encoder architecture, while ``m6'' uses a single encoder. Table~\ref{tab:encoder_design_abl} presents a controlled comparison where both models use a channel attention fusion mechanism. The ``m6'' model, despite having significantly fewer parameters (64.3K vs. 118.5K), consistently outperforms the larger dual-encoder ``m5'' model. This result was pivotal, leading us to abandon the more complex dual-encoder structure. We concluded that allocating parameters towards a more effective fusion and body in a single-encoder framework provides a better performance-efficiency trade-off, which became a core principle in designing the final ``PanTiny'' model.

\begin{table*}[h!]
\centering
\resizebox{\textwidth}{!}{%
\begin{tabular}{l|c|ccc|ccc|ccc}
\toprule
\multirow{2}{*}{\textbf{Model (Encoder Type)}} & \multirow{2}{*}{\textbf{Params(K)}} & \multicolumn{3}{c}{\textbf{WV2}} & \multicolumn{3}{c}{\textbf{WV3}} & \multicolumn{3}{c}{\textbf{GF2}} \\
\cmidrule(lr){3-5} \cmidrule(lr){6-8} \cmidrule(lr){9-11}
& & \textbf{PSNR}$\uparrow$ & \textbf{SSIM}$\uparrow$ & \textbf{SAM}$\downarrow$ & \textbf{PSNR}$\uparrow$ & \textbf{SSIM}$\uparrow$ & \textbf{SAM}$\downarrow$ & \textbf{PSNR}$\uparrow$ & \textbf{SSIM}$\uparrow$ & \textbf{SAM}$\downarrow$ \\
\midrule
m5 (Dual-Encoder, Large) & 118.5 & 41.05 & \textbf{0.9642} & \textbf{0.0246} & 29.89 & 0.9119 & 0.0842 & 47.45 & 0.9860 & 0.0109 \\
\textbf{m6 (Single-Encoder)} & \textbf{64.3} & \textbf{41.06} & 0.9641 & 0.0248 & \textbf{30.00} & \textbf{0.9133} & \textbf{0.0830} & \textbf{47.54} & \textbf{0.9864} & \textbf{0.0108} \\
\bottomrule
\end{tabular}%
}
\caption{Comparison between our single-encoder (``m6'') and dual-encoder (``m5'') experimental models. The single-encoder design achieves superior performance with fewer parameters.}
\label{tab:encoder_design_abl}
\end{table*}

\subsection{Full Ablation Results for Final Model Components}
The main paper presented condensed versions of our final fusion and refinement ablations for brevity. Here, we provide the complete tables with all metrics (Table~\ref{tab:fusion_full} and Table~\ref{tab:refine_full}). These results reinforce our conclusion that for ``PanTiny'', simple and well-chosen convolutional blocks outperform more complex alternatives in the multi-dataset setting, providing the best balance of parameter efficiency and performance.

\begin{table*}[h!]
\centering
\resizebox{\textwidth}{!}{%
\begin{tabular}{l|c|ccc|ccc|ccc}
\toprule
\multirow{2}{*}{\textbf{Fusion Type}} & \textbf{Params} & \multicolumn{3}{c}{\textbf{WV2}} & \multicolumn{3}{c}{\textbf{WV3}} & \multicolumn{3}{c}{\textbf{GF2}} \\
\cmidrule(lr){3-5} \cmidrule(lr){6-8} \cmidrule(lr){9-11}
& \textbf{(K)} & \textbf{PSNR}$\uparrow$ & \textbf{SSIM}$\uparrow$ & \textbf{SAM}$\downarrow$ & \textbf{PSNR}$\uparrow$ & \textbf{SSIM}$\uparrow$ & \textbf{SAM}$\downarrow$ & \textbf{PSNR}$\uparrow$ & \textbf{SSIM}$\uparrow$ & \textbf{SAM}$\downarrow$ \\
\midrule
1x1 Conv & 68.2 & \underline{41.75} & \underline{0.9690} & \underline{0.0227} & \underline{30.45} & \underline{0.9222} & \underline{0.0761} & \underline{48.37} & \underline{0.9888} & 0.0098 \\
Channel Attn. & 71.4 & 41.72 & 0.9686 & 0.0228 & 30.44 & 0.9216 & 0.0767 & 48.34 & 0.9886 & \underline{0.0097} \\
Gated Conv & 70.3 & 41.66 & 0.9686 & 0.0229 & 30.44 & 0.9219 & 0.0766 & 48.32 & 0.9886 & 0.0098 \\
DeepFusion \cite{panmamba} & 113.6 & 41.66 & 0.9684 & 0.0229 & 30.34 & 0.9206 & 0.0771 & 48.35 & 0.9887 & 0.0098 \\
\textbf{Enhanced Conv (Ours)} & \textbf{81.7} & \textbf{41.85} & \textbf{0.9696} & \textbf{0.0224} & \textbf{30.59} & \textbf{0.9238} & \textbf{0.0749} & \textbf{48.61} & \textbf{0.9894} & \textbf{0.0095} \\
\bottomrule
\end{tabular}%
}
\caption{Full ablation results for the fusion module in the final ``PanTiny'' architecture. Our ``Enhanced Conv'' provides the best overall trade-off.}
\label{tab:fusion_full}
\end{table*}

\begin{table*}[h!]
\centering
\resizebox{\textwidth}{!}{%
\begin{tabular}{l|c|ccc|ccc|ccc}
\toprule
\multirow{2}{*}{\textbf{Refine Type}} & \textbf{Params} & \multicolumn{3}{c}{\textbf{WV2}} & \multicolumn{3}{c}{\textbf{WV3}} & \multicolumn{3}{c}{\textbf{GF2}} \\
\cmidrule(lr){3-5} \cmidrule(lr){6-8} \cmidrule(lr){9-11}
& \textbf{(K)} & \textbf{PSNR}$\uparrow$ & \textbf{SSIM}$\uparrow$ & \textbf{SAM}$\downarrow$ & \textbf{PSNR}$\uparrow$ & \textbf{SSIM}$\uparrow$ & \textbf{SAM}$\downarrow$ & \textbf{PSNR}$\uparrow$ & \textbf{SSIM}$\uparrow$ & \textbf{SAM}$\downarrow$ \\
\midrule
\textbf{Conv (Ours)} & \textbf{81.7} & \textbf{41.90} & \underline{0.9697} & \underline{0.0224} & \textbf{30.61} & \textbf{0.9240} & \textbf{0.0749} & 48.49 & \textbf{0.9891} & 0.0097 \\
Channel Attn. & 96.4 & \textbf{41.90} & \textbf{0.9698} & \textbf{0.0223} & \underline{30.55} & \underline{0.9230} & \underline{0.0751} & \underline{48.50} & \textbf{0.9891} & \textbf{0.0096} \\
Large Conv & 88.8 & \underline{41.87} & 0.9696 & \underline{0.0224} & 30.49 & 0.9225 & 0.0759 & \textbf{48.52} & \textbf{0.9891} & \textbf{0.0096} \\
\bottomrule
\end{tabular}%
}
\caption{Full ablation results for the refinement module in the final ``PanTiny'' architecture. A simple convolution is most effective.}
\label{tab:refine_full}
\end{table*}

\section{Additional Visual Results}
\label{sec:supp_visuals}
To save space in the main paper, we presented a limited set of visual comparisons. This section provides additional qualitative examples to complement the quantitative results. These examples offer a more intuitive understanding of the performance differences between various methods across all three datasets (WV2, WV3, and GF2) and demonstrate the robustness of our approach.

\begin{figure}[h!]
\centering
\includegraphics[width=\linewidth]{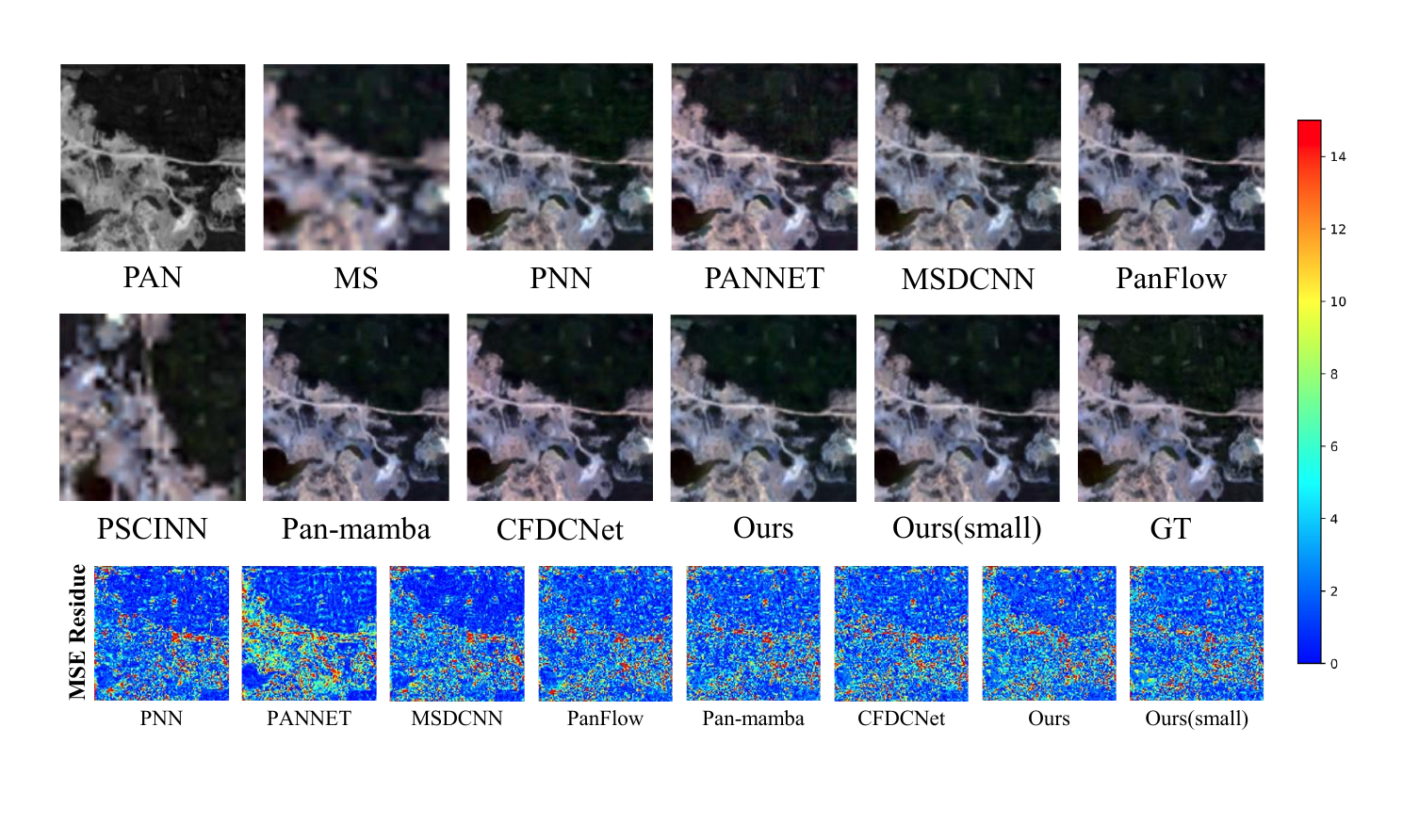}
\caption{Visual comparison on the WorldView-2 (WV2) dataset.}
\label{fig:vp_wv2}
\end{figure}

\begin{figure}[h!]
\centering
\includegraphics[width=\linewidth]{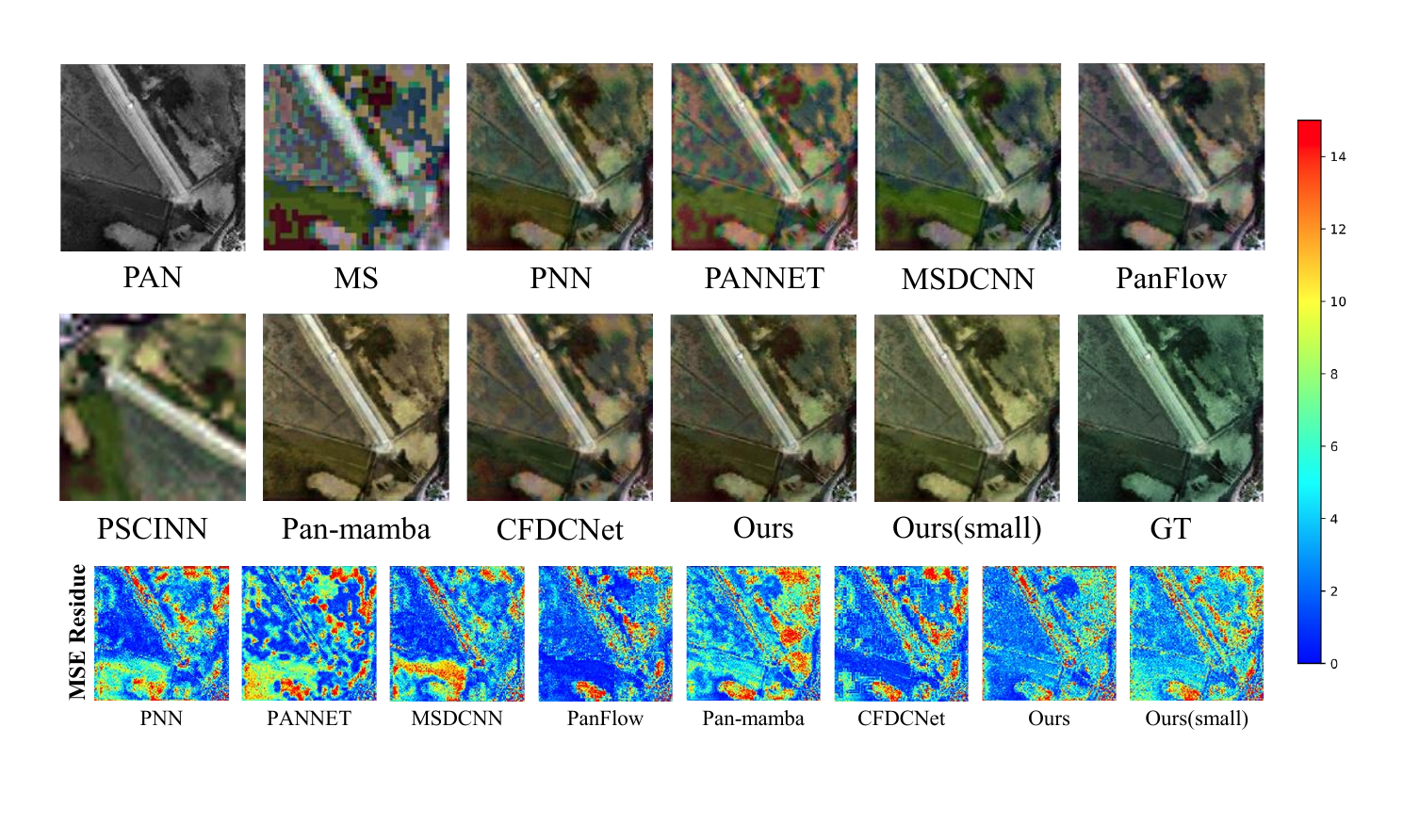}
\caption{Visual comparison on the WorldView-3 (WV3) dataset. Our method performs exceptionally well when the multispectral (MS) image contains a significant amount of noise.}
\label{fig:vp_wv3}
\end{figure}

\begin{figure}[h!]
\centering
\includegraphics[width=\linewidth]{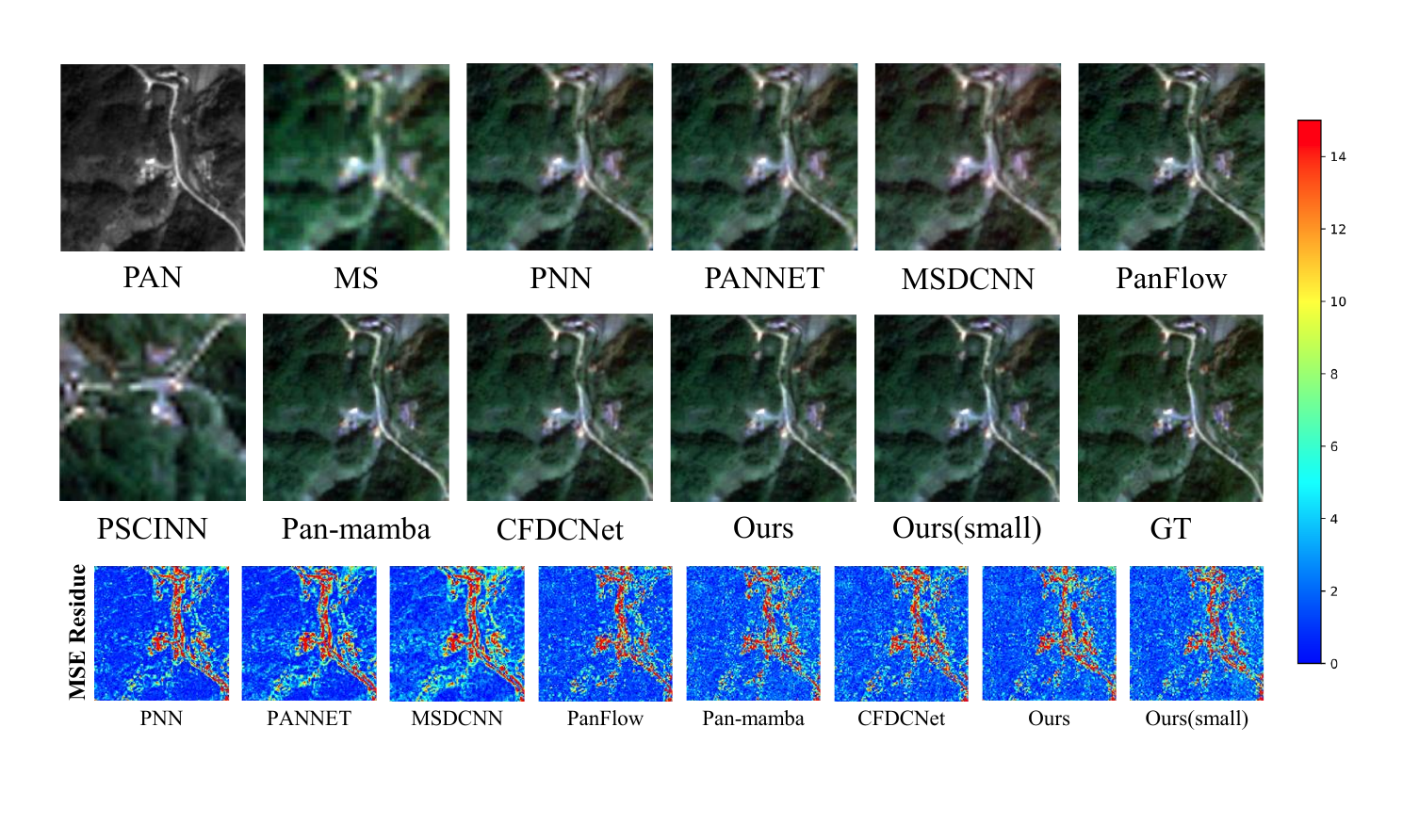}
\caption{Visual comparison on the GaoFen-2 (GF2) dataset.}
\label{fig:vp_gf2}
\end{figure}

\section{Codebase and Reproducibility}
\label{sec:supp_code}
To ensure full reproducibility and facilitate future research, we provide a comprehensive and easy-to-use codebase as a ``code.tar.gz'' archive in the supplementary materials. Our framework is built around a unified experiment runner that leverages a hierarchical YAML configuration system. This allows researchers to define a base configuration and then specify a series of experiments that inherit and override these settings, enabling efficient and organized ablation studies. For more details, please refer to the ``README.md'' file included in our supplementary materials.



\end{document}